\pdfoutput=1
\documentclass[11pt]{article}

\PassOptionsToPackage{pagebackref}{hyperref}
\usepackage[preprint]{acl}
\usepackage[T1]{fontenc}
\usepackage[utf8]{inputenc}
\usepackage[english]{babel}
\usepackage{amsmath}
\usepackage{mathtools} 
\usepackage{newtxtext}   
\usepackage{newtxmath}   
\usepackage{hyperref}
\usepackage{microtype}
\usepackage{inconsolata}
\usepackage{booktabs}
\usepackage{multirow}
\usepackage{graphicx}
\usepackage{scalerel}
\usepackage{caption}
\usepackage{float}
\usepackage{titlesec}
\usepackage{enumitem}
\usepackage{tikz}
\usepackage{colortbl}   
\usepackage{pgf}        
\usepackage{dblfloatfix}
\usepackage[noabbrev,capitalize,nameinlink]{cleveref}
\usepackage{comment}

\hypersetup{
    pdftitle={OctoLong: Mid-Training On Cross-Repository Code Contexts Enhances Long-Context Modeling}
}
\makeatletter
\ifacl@finalcopy
    \hypersetup{pdfauthor={Indraneil Paul, Falko Helm, Goran Glava\v{s}, Iryna Gurevych}}
\fi
\makeatother


\definecolor{CommentGrey}{HTML}{949494}
\definecolor{DeepForest}{HTML}{227805}
\definecolor{DeepBlood}{HTML}{780505}
\definecolor{Twitch}{HTML}{8c34eb}
\definecolor{Saffron}{HTML}{eb5c34}
\definecolor{Navy}{HTML}{2c59a5}
\newcommand{\diffup}[1]{\small{\textbf{\color{DeepForest}\texttt{+#1}}}}
\newcommand{\diffdo}[1]{\small{\textbf{\color{DeepBlood}\texttt{-#1}}}}

\AtBeginDocument{%
}
\makeatletter
\@ifpackageloaded{backref}{%
}{}
\makeatother

\DeclareRobustCommand{\numcircle}[1]{%
  \tikz[baseline=(char.base)]{
    \node[shape=rectangle, rounded corners=2pt, fill=black, text=white,
          inner sep=0pt, minimum size=1em, font=\footnotesize\bfseries] (char) {#1};
  }%
}

\definecolor{OctoBlue}{HTML}{2c59a5}   
\definecolor{translateBlue}{HTML}{4C8BF5}   
\newcommand{\gradientcell}[6]{%
    \ifdimcomp{#1pt}{>}{#3pt}{\cellcolor{#5!100.0!#4!#6}\texttt{#1}}{%
        \ifdimcomp{#1pt}{<}{#2pt}{\cellcolor{#5!0.0!#4!#6}\texttt{#1}}{%
            \pgfmathparse{int(round(100*(#1-#2)/(#3-#2)))}%
            \xdef\tempa{\pgfmathresult}%
            \cellcolor{#5!\tempa!#4!#6}\texttt{#1}%
        }}%
}
\newcommand{\olsc}[1]{\gradientcell{#1}{0}{80}{white}{translateBlue}{70}}
\newcommand{\ollcr}[1]{\gradientcell{#1}{0}{25}{white}{translateBlue}{70}}
\newcommand{\olcg}[1]{\gradientcell{#1}{0}{55}{white}{translateBlue}{70}}
\providecommand{\hficon}{\raisebox{-1.5pt}{\includegraphics[height=9pt]{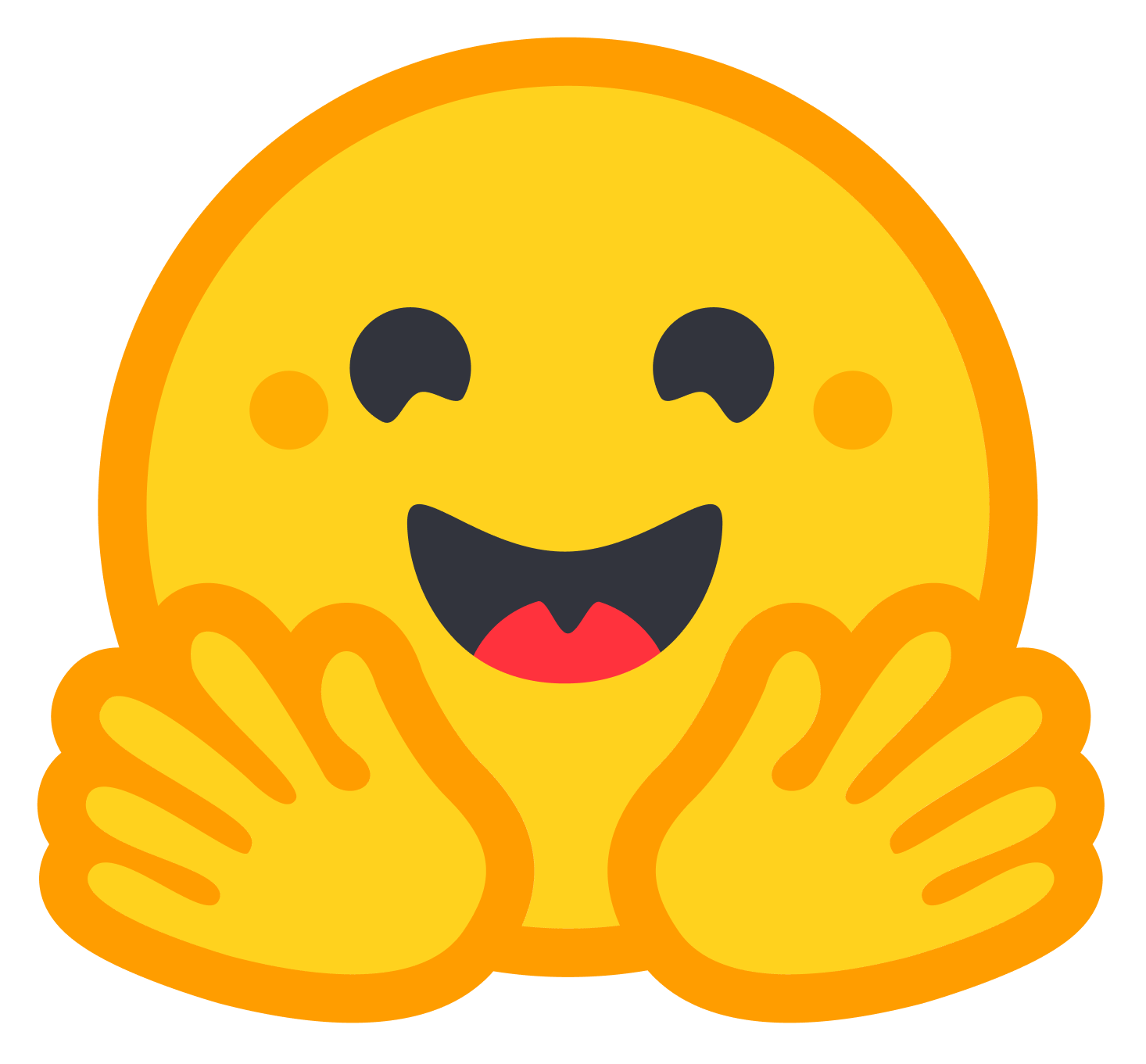}}}
\providecommand{\giticon}{\raisebox{-1.5pt}{\includegraphics[height=9pt]{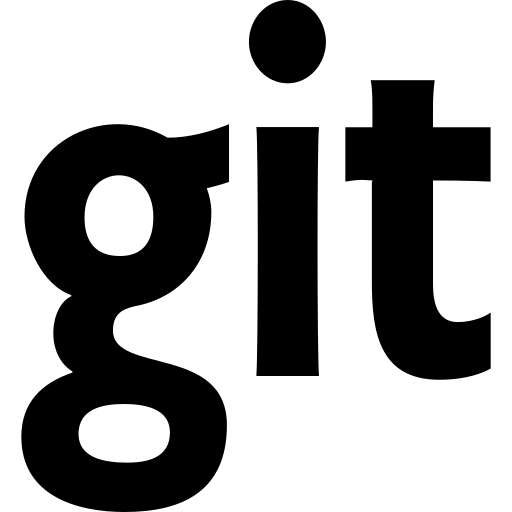}}}
\newcommand{\gitrepo}[1]{\giticon\;\href{https://github.com/#1}{\texttt{#1}}}
\providecommand{\hfmodel}[1]{\hficon\;\href{https://huggingface.co/#1}{\texttt{\detokenize{#1}}}}
\newcommand{\hsq}[1]{\scalebox{0.9}[1]{#1}}
\newcommand{\rmodel}[1]{\hficon\;\href{https://huggingface.co/#1}{\textcolor{black}{\hsq{\texttt{\detokenize{#1}}}}}}
\newcommand{\gcode}{\inlineicon{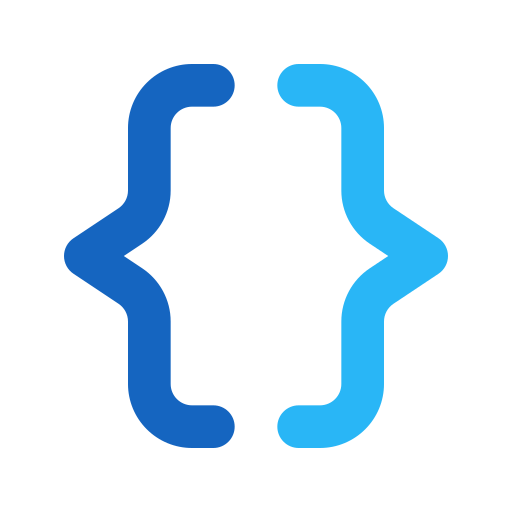}}
\newcommand{\gpost}{\inlineicon{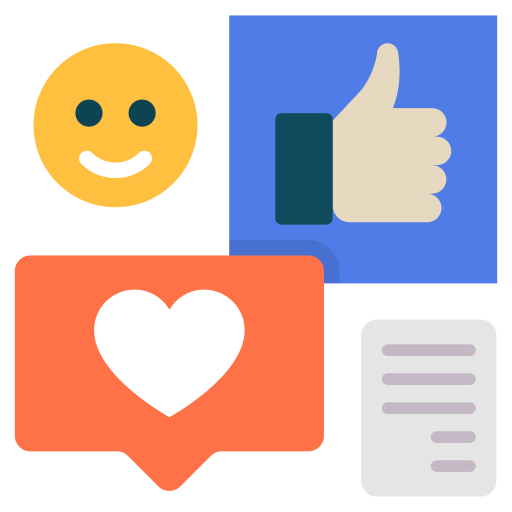}}
\newcommand{\gcext}{\inlineicon{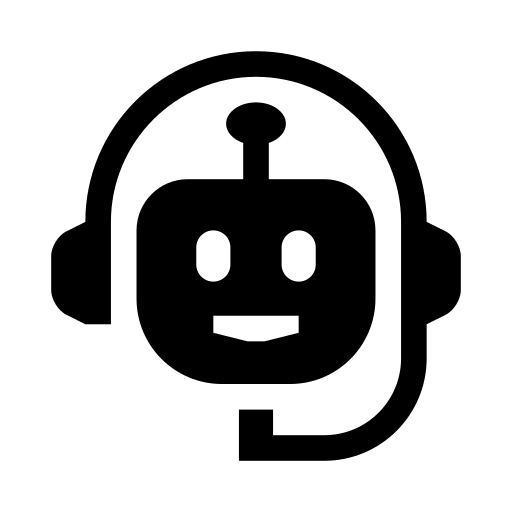}}

\newcommand{\rparagraph}[1]{\vspace{1.0mm}\noindent\textbf{#1.}}
\newcommand{\rrparagraph}[1]{\vspace{0.6mm}\noindent\textbf{\textsc{#1:}}}

\newcommand{\hsep}{\newline}
\makeatletter
\def\sqsplit@#1{%
  \ifx#1\@nil\else
    \scalebox{0.85}[1]{#1}\penalty\z@\hskip\z@\relax
    \expandafter\sqsplit@
  \fi}
\newcommand{\sqsplit}[1]{\sqsplit@#1\@nil}
\newcommand{\sqsplitds}[1]{\expandafter\sqsplit\expandafter{\detokenize{#1}}}
\makeatother
\newcommand{\hfds}[1]{\hficon\;\href{https://huggingface.co/datasets/#1}{\textcolor{black}{\texttt{\sqsplitds{#1}}}}}
\newcommand{\dataset}[4]{%
  \item \makebox[\linewidth]{\textbf{\texttt{#1}}\hfill{\small\textbf{\texttt{#3~SAMPLES}}}}%
    \ifnotempty{#2}{\par\vspace{0.2ex}{\raggedright\small #2\par}}%
    \par
  #4%
}
\newcommand{\dscatglyph}[1]{\raisebox{-1.5pt}{\includegraphics[height=11pt]{#1}}}
\newcommand{\dscat}[2]{%
  \par\vspace{1.6mm}\noindent
  \dscatglyph{#1}\;\textbf{\texttt{#2}}\par\vspace{0.3mm}%
}

\makeatletter
\long\def\@makefntext#1{%
  \fontsize{8.1}{9}\selectfont
  \parindent 1em%
  \noindent
  \hb@xt@1.8em{\hss\@makefnmark}#1}
\makeatother

\newlength{\iconht}
\newcommand{\inlineicon}[1]{\raisebox{-1.5pt}{\includegraphics[height=\iconht]{#1}}}

\newcommand{\upbetter}{\raisebox{-0.15ex}{\includegraphics[height=1.3ex]{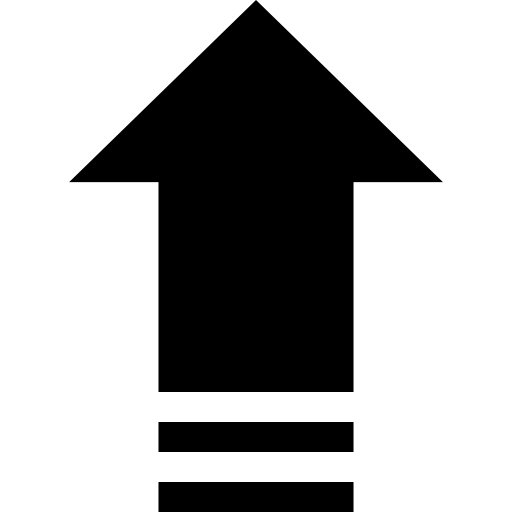}}}
\newcommand{\downbetter}{\raisebox{-0.15ex}{\includegraphics[height=1.3ex]{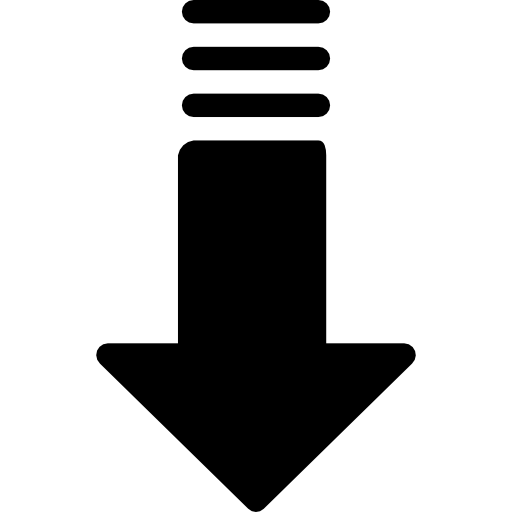}}}
\newcommand{\pmstd}[1]{\,{\small\textbf{\texttt{{\boldmath$\pm$}\,#1}}}}
\newcommand{\hc}[1]{\multicolumn{1}{c}{#1}}

\title{%
\makebox[0pt][r]{\raisebox{-0.28\height}{\includegraphics[height=1.45\baselineskip]{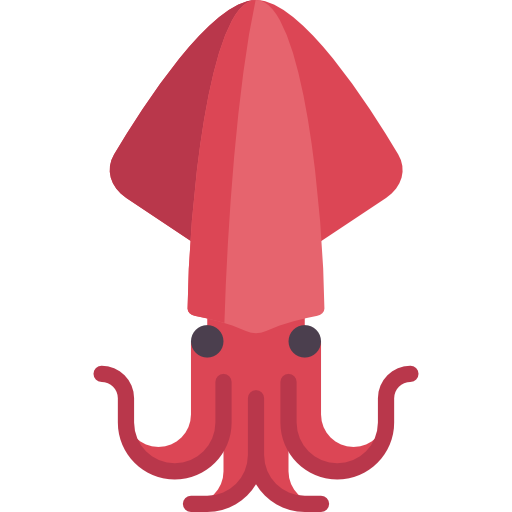}}\hspace{0.5em}}%
\begin{tabular}[c]{@{}c@{}}%
\texttt{OctoLong}: Mid-Training On Cross-Repository Code\\
Contexts Enhances Long-Context Modeling%
\end{tabular}%
}

\author{
Indraneil Paul\textsuperscript{1},
Falko Helm\textsuperscript{1},
Goran Glava\v{s}\textsuperscript{2},
\and
Iryna Gurevych\textsuperscript{1}
\\
\textsuperscript{1} UKP Lab \& Center for Applied Cybersecurity ATHENE, TU Darmstadt\\
\textsuperscript{2} CAIDAS, University of Würzburg \\
\vspace{-1.8em} \\
}

\makeatletter
\AddToHook{cmd/appendix/before}{\def\cref@section@alias{appendix}}
\newcommand{\ifnotempty}[2]{%
  \def\@tempa{#1}\ifx\@tempa\empty\else#2\fi
}
\makeatother

\newcounter{rqcnt}
\renewcommand{\therqcnt}{\arabic{rqcnt}}
\crefname{rqcnt}{RQ}{RQs}
\Crefname{rqcnt}{RQ}{RQs}
\crefformat{rqcnt}{#2\texttt{RQ#1}#3}
\Crefformat{rqcnt}{#2\texttt{RQ#1}#3}
\newcommand{\rqhead}[2]{%
  \par\vspace{1.4mm}\noindent
  \refstepcounter{rqcnt}\label{#1}%
  \textbf{\texttt{\large RQ\therqcnt:} \large #2}\par
}

\ifdefined\ThesisMode\AtBeginDocument{\ifdefined\ThesisPageOffset\setcounter{page}{\ThesisPageOffset}\fi}\fi

\begin{document}
\maketitle
\begin{abstract}

Context lengths of language models (LMs) have dramatically increased, driven by the demands for in-context learning, self-improvement, and long-horizon agentic workflows. 
However, existing long-context corpora are dominated by books, academic articles, and code repositories, which are finite and often lack long-distance dependencies. 
In this work, we introduce \textbf{\texttt{OctoLong}}, a context-engineering pipeline that instruments an AST parser, a language server backend, and a package manager to recursively retrieve code references, enabling the curation of dependency-rich code contexts up to millions of tokens long. We then train \textbf{\texttt{OctoLong-Instruct}}, a suite of capable long-context open LMs, derived from base models ranging in size from 600M to 14B parameters, via context-extension mid-training on a $\approx$50B-token mixture containing $\approx$6.2B tokens of \texttt{OctoLong} code contexts, followed by $\approx$10B tokens of instruction tuning. 
Our ablations and experimental evaluations against 18 strong long-context open-weight LMs show that supplanting just 12\% of conventional context-extension corpora with \texttt{OctoLong} data yields substantial gains in long-range retrieval, long-term state tracking, repository-level code understanding, and downstream agentic tasks, while also enhancing API usage in regular coding scenarios.

\end{abstract}

\section{Introduction}
\label{sec:Intro}
\begin{figure}
    \centering
    \includegraphics[width=\linewidth,height=4.3cm]{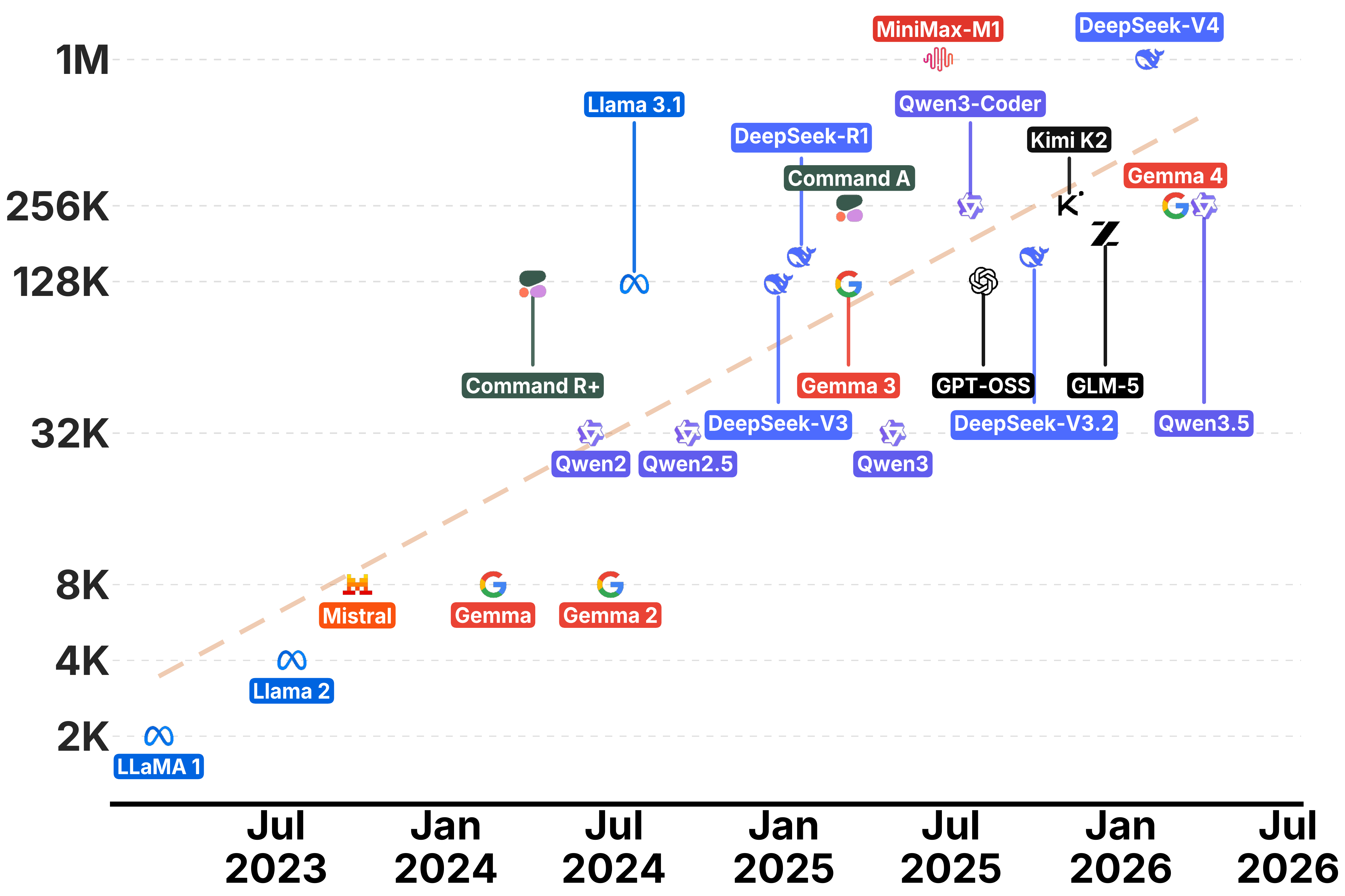}

    \caption{Context lengths of open-weight LMs. The trend is exponential, doubling on average every 5 months.}
    \label{fig:ContextTrend}
    \vspace{-1.5em}
\end{figure}

\vspace{-0.25em}
Emergent language model (LM) applications such as in-context learning~\citep{DBLP:conf/naacl/BertschIXABGN25}, retrieval-augmented generation~\citep{DBLP:conf/www/Qian0ZMLD025, DBLP:journals/corr/abs-2502-06049}, and agentic tool-use~\citep{DBLP:journals/corr/abs-2601-11868, DBLP:conf/iclr/JimenezYWYPPN24} demand long-horizon capabilities~\citep{DBLP:conf/nips/KwaWBDGHJKRABBD25}, exponentially driving up the demand on the context lengths the LMs are expected to support (ref. \cref{fig:ContextTrend}). Prior attempts to extend the context length mainly involve architectural interventions such as sparse attention~\citep{DBLP:conf/acl/YuanGD0ZZXWW0WR25, DBLP:conf/naacl/HanWPX0JW24, DBLP:journals/corr/abs-2004-05150} and linear attention~\citep{DBLP:journals/corr/abs-2605-22791, DBLP:conf/icml/DaoG24}. In isolation, these approaches deteriorate the performance on longer contexts~\citep{DBLP:journals/corr/abs-2409-12181, DBLP:journals/corr/abs-2407-08112}, causing LM developers to turn to orthogonal axes of improvement such as hardware-use improvements~\citep{DBLP:journals/corr/abs-2310-03294, DBLP:journals/corr/abs-2310-01889}, inference-time context management~\citep{DBLP:conf/nips/LiHYVLYCLC24, DBLP:conf/nips/Zhang00CZC0TRBW23}, and better data engineering~\citep{DBLP:journals/corr/abs-2502-16684}.

\rparagraph{Data Engineering for Context Extension} Data engineering for long-context training~\citep{DBLP:conf/icml/FuPNYHK024} focuses on engineering long text inputs with a high density of referrals (i.e., dependencies) spanning a range of distances. Existing approaches span the gamut of token-level dependency filtering~\citep{DBLP:journals/corr/abs-2510-25804, DBLP:journals/corr/abs-2601-21571}, curating multiple salient documents in context~\citep{DBLP:conf/icml/Gao0LZ025, DBLP:journals/corr/abs-2510-02330, DBLP:journals/corr/abs-2411-08147}, and referral-based packing~\citep{DBLP:conf/iclr/0001HY00X025, DBLP:conf/aaai/StaniszewskiTJ025}. Practically, however, the contexts engineered by such approaches often lack complex multi-hop dependencies~\citep{DBLP:conf/acl/LiLL024} and mine irrelevant or contradictory information~\citep{DBLP:conf/emnlp/DuTRRBGWSHP25, DBLP:journals/corr/abs-2510-05862}. 
In this work, we introduce \textit{cross-repository code dependency mining}. This referral packing technique taps into openly available code data to unlock a virtually limitless source of multi-hop referral-dense contexts that transcend repository boundaries.

\rparagraph{Contributions} We address the abovementioned research gaps and make the following contributions:

\vspace{0.15em}

\noindent \numcircle{1} We introduce \texttt{OctoLong}, a pipeline to generate cross-repository code contexts using an AST parser, an LSP server, and a package manager in tandem. Analyses of attention maps and model likelihoods show that our pipeline produces exceptionally dependency-rich long contexts (\cref{sec:Dataset}). 

\vspace{0.15em}

\noindent \numcircle{2} We collect $\approx$6.2B tokens of cross-repository long-context data with the \texttt{OctoLong} pipeline and combine it with other long-data sources to curate the $\approx$50B-token \texttt{OctoLong-LCFT} corpus. We then compile the $\approx$10B-token \texttt{OctoLong-SFT} mix of open-source instruction and agentic data (\cref{sec:Setup}). 

\vspace{0.15em}

\noindent \numcircle{3} Using the above corpora, we subject Qwen3 dense base models to long-context fine-tuning (LCFT), quadrupling their supported context length to 128K, followed by supervised fine-tuning (SFT) to obtain the \texttt{OctoLong-Instruct} suite (\cref{sec:Setup}).

\vspace{0.15em}

\noindent \numcircle{4} We establish, via extensive experiments and ablations, that LCFT with our \texttt{OctoLong} data in the long-context data mix leads to substantial improvements not only in long-context code tasks but also in general-domain long-context applications including agentic use-cases,  without penalty on short-context LM capabilities (\cref{sec:Results}).

\section{Related Work}
\label{sec:Related_Work}
\vspace{-0.25em}
We briefly outline two threads of relevant existing work: \numcircle{1} cross-document context engineering and \numcircle{2} long-context code data mining.

\rparagraph{Cross-Document Context Engineering} Prior work towards engineering cross-document contexts has relied on hyperlinks~\citep{DBLP:conf/iclr/0001HY00X025, DBLP:conf/acl/YasunagaLL22}, book BISAC codes~\citep{DBLP:conf/aaai/JiaWGCLLWXJZG26}, event co-references~\citep{DBLP:conf/emnlp/CaciularuCBPCD21}, and repository-level code dependencies~\citep{DBLP:journals/corr/abs-2401-14196}. These approaches are all either hard to scale or yield loosely related contexts. In contrast, we detail how to recursively mine referred class and function implementations beyond repository boundaries --- an untapped source of tightly-coupled data whose acquisition is easy to scale.

\rparagraph{Mining Long Self-Contained Code Data} Early file-level attempts to mine long code data found that long code files are usually low-quality configuration files and uninformative build artifacts~\citep{DBLP:journals/corr/abs-2407-00434}. Subsequent methods have since moved on to serializing code at the repository level~\citep{DBLP:conf/coling/DingWARNBRX24, DBLP:journals/corr/abs-2406-01359}. Serialized repositories have since become the anchor of mid-training~\citep{DBLP:journals/corr/abs-2510-14865, DBLP:journals/corr/abs-2406-03476} and context-extension~\citep{DBLP:journals/corr/abs-2510-06826} corpora. However, most code is not self-contained at the repository level~\citep{DBLP:conf/icse/YuSRZZMLLWX24}, which restricts models' ability to reason about multi-repository projects~\citep{DBLP:journals/corr/abs-2603-03194}. Attempts to train on cross-repository contexts are stymied by inadequate tooling~\citep{DBLP:journals/corr/abs-2601-10773} and have thus been limited to sourcing dependency library names~\citep{DBLP:journals/tse/LiaoPSRHXJL24}. In \cref{sec:Dataset}, we detail a novel pipeline to source referred implementations beyond repository bounds.

\section{The \texorpdfstring{\texttt{OctoLong}}{OctoLong} Dataset}
\label{sec:Dataset}
\vspace{-0.25em}
Sourcing long-context code for language modeling has traditionally focused on serializing \textit{individual} code repositories in dependency traversal order~\citep{DBLP:journals/corr/abs-2402-19173, DBLP:journals/corr/abs-2401-14196, DBLP:conf/emnlp/ZhangCZKLZMLC23, DBLP:journals/corr/abs-2306-10998, DBLP:journals/corr/abs-2510-13697}. To test our hypotheses from \cref{sec:Intro}, we develop the \texttt{OctoLong} pipeline to retrieve code dependencies \textit{across} repository boundaries.

\rparagraph{Language Server Protocol} We leverage the Language Server Protocol (LSP),\footnote{\href{https://microsoft.github.io/language-server-protocol/}{\scalerel*{\includegraphics{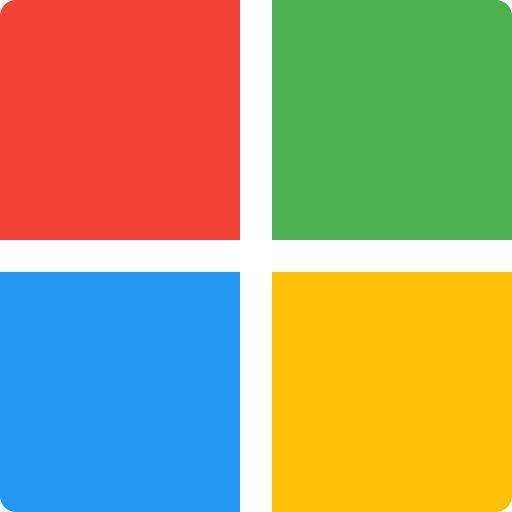}}{|} \texttt{Language Server Protocol}}} a JSON-RPC standard that provides a unified language- and client-agnostic wire interface to source code analyses by fronting language-specific compilers. Crucial for our purposes, mature language server implementations for popular programming languages can use the established conventions of detected package managers, in tandem with the abstract syntax tree (AST) information produced by the compiler, to map the location of imported classes and functions. The LSP ecosystem, traditionally used by IDEs, offers a robust, type- and scope-aware approach to retrieving dependency code that predecessors reliant on pure syntactic parsing cannot achieve.\footnote{\gitrepo{universal-ctags/ctags}}

\begin{figure*}[t!]
    \centering
    \includegraphics[width=\linewidth, height=4.4cm]{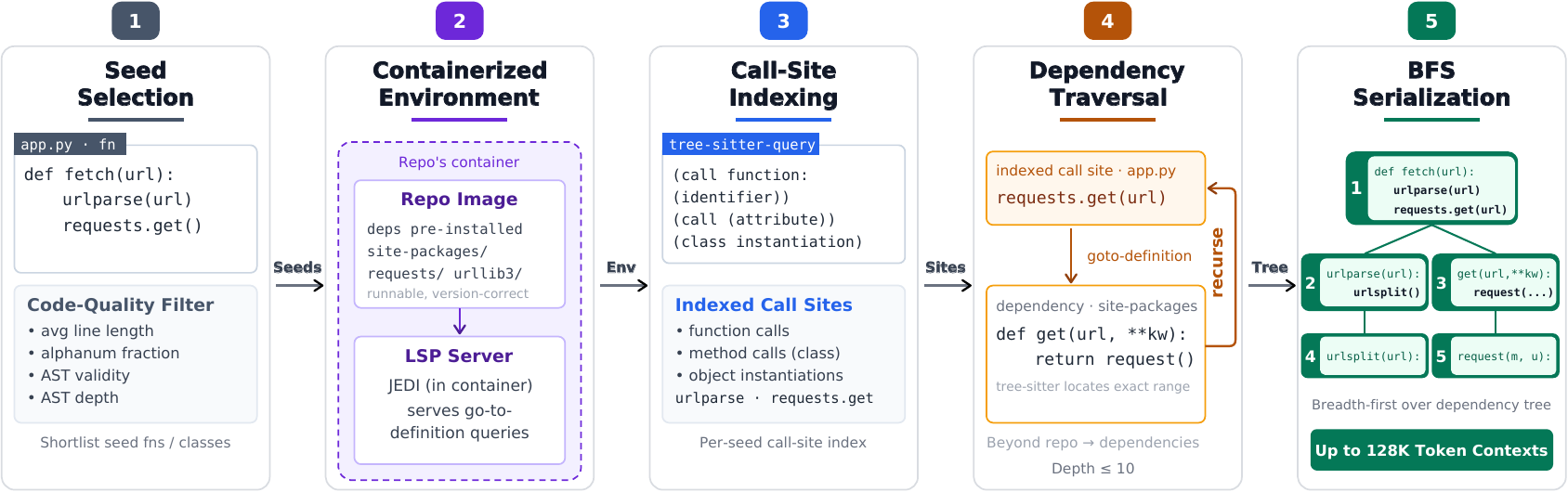}
    \caption{Overview of the \texttt{OctoLong} pipeline's core stages. Utilizing AST information, a language server backend leverages package manager metadata to recursively retrieve dependency data and engineer very long code contexts. BFS hops are prepended, ensuring the serialized context runs \textcolor{DeepForest}{\texttt{[deepest deps $\to$ direct deps $\to$ seed]}}.}
    \label{fig:OctoLong}
    \vspace{-1em}
\end{figure*}

\rparagraph{\texttt{OctoLong} Data Acquisition} \cref{fig:OctoLong} illustrates all stages of the \texttt{OctoLong} data acquisition pipeline for long code contexts. We start by sourcing from SEART Hub~\citep{DBLP:conf/icsm/DabicTB24} reputable and actively maintained Python repositories\footnote{While our data collection is restricted to Python, the \texttt{OctoLong} pipeline is extensible to any programming language with a mature LSP backend and a package manager.}, with at least 15 GitHub stars and more than 5 contributors. From these, we retain only those that (a) declare dependencies and (b) are installable, resulting in 12K repositories. Each Python file from retained repositories becomes a candidate seed: we keep the 92K seeds that pass OpenCoder's default code-quality filtering pipeline~\citep{DBLP:conf/acl/HuangCLXHSXYLZC25}. Subsequently, we provision each seed in a containerized sandbox with its repository and dependency environment and traverse the dependency graph via breadth-first search (BFS). We traverse up to a pre-specified depth of 10 steps or when we collect 128K tokens of code (whichever is fulfilled first), prepending the code retrieved in each step to the context. 
Each hop consists of a \texttt{tree-sitter}\footnote{\gitrepo{tree-sitter/tree-sitter}} query that extracts class and function call-sites, followed by a \texttt{GoToDeclaration} call\footnote{While the LSP standard defines a \texttt{GoToImplementation} endpoint, which would significantly simplify our collection, this remains unsupported by most language servers.} per call-site using the \texttt{jedi}\footnote{\gitrepo{pappasam/jedi-language-server}} language server; this is followed by another \texttt{tree-sitter} query that obtains the function or class boundaries and extracts the implementation from the target file. 
We ensure we never repeat classes or functions. We also discard contexts shallower than 5 levels as well as those with a hop success rate below 60\%. Following this procedure, we end up with $\approx$6.2B code tokens. 

\begin{table*}[t!]
    \centering
    \scalebox{0.8}{\footnotesize \upbetter\;\textbf{\texttt{Higher is Better}} \qquad \downbetter\;\textbf{\texttt{Lower is Better}}}
    \scalebox{0.55}{
    \setlength{\tabcolsep}{6pt}
    \begin{tabular}{@{}ll rrrr rrr c@{}}
        \toprule

        \multirow{3}{*}{\scalerel*{\includegraphics{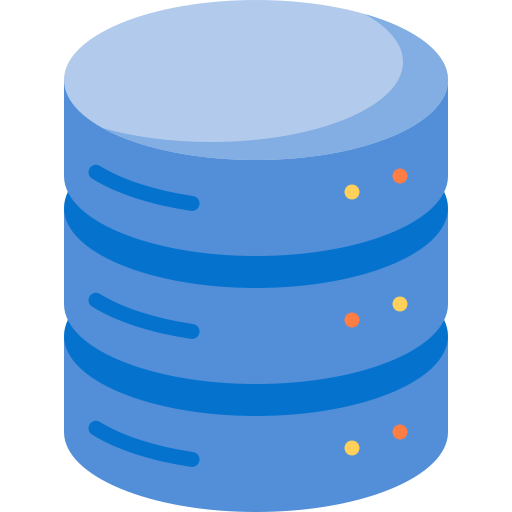}}{B}\;\textbf{\texttt{Source}}}
        & \multirow{3}{*}{\shortstack[l]{\textbf{\texttt{Replication}}\\ \textbf{\texttt{Source}}}}
        & \multicolumn{4}{c}{\multirow{2}{*}{\textbf{\texttt{Length Distribution Percentile}}}}
        & \multicolumn{3}{c}{\textbf{\texttt{Dependency Score}}}
        & \multirow{3}{*}{\shortstack{\textbf{\texttt{Estimated}}\\ \textbf{\texttt{Open Data}}}} \\
        \cmidrule(lr){7-9}

        & & \multicolumn{4}{c}{}
        & \multicolumn{2}{c}{\textbf{\texttt{LongAttn}}} & \textbf{\texttt{LongPPL}} & \\
        \cmidrule(lr){3-6} \cmidrule(lr){7-8} \cmidrule(lr){9-9}

        & & \texttt{50th} & \texttt{90th} & \texttt{95th} & \texttt{99th}
        & \shortstack[r]{\texttt{Dep.}\\ \texttt{Mean}\;\upbetter}
        & \shortstack[r]{\texttt{Sink Prop.}\\ \texttt{(Pos 0--32)}\;\downbetter}
        & \shortstack[r]{\texttt{\% Key}\\ \texttt{Tokens}\;\upbetter} & \\

        \midrule

        \multirow{2}{*}{\scalerel*{\includegraphics{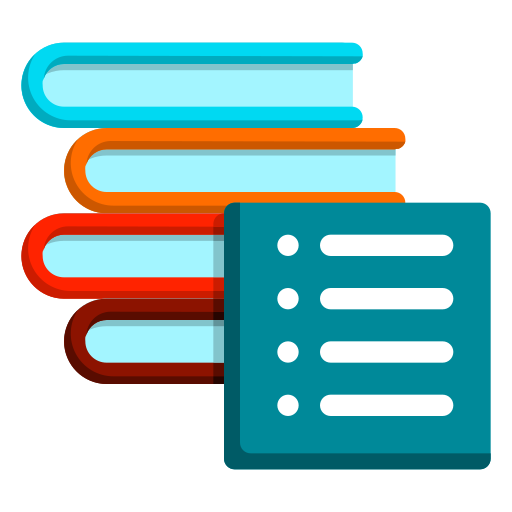}}{B}\;\texttt{Books}} & \texttt{RedPajama}
        & \multirow{2}{*}{\texttt{88,329}} & \multirow{2}{*}{\texttt{221,456}} & \multirow{2}{*}{\texttt{288,375}} & \multirow{2}{*}{\texttt{453,776}}
        & \multirow{2}{*}{\texttt{0.5581}} & \multirow{2}{*}{\texttt{6.1e-4}} & \multirow{2}{*}{\texttt{0.87}} & \multirow{2}{*}{\shortstack{\texttt{< 1T}\\ \texttt{Tokens}}} \\
        & \citep{DBLP:conf/nips/WeberFAOAALNYAA24} & & & & & & & & \\

        \addlinespace

        \multirow{2}{*}{\scalerel*{\includegraphics{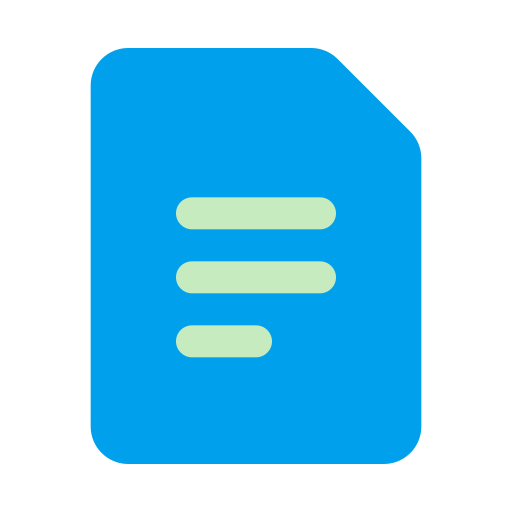}}{B}\;\texttt{Academic Papers}} & \texttt{Common Pile}
        & \multirow{2}{*}{\texttt{16,565}} & \multirow{2}{*}{\texttt{40,387}} & \multirow{2}{*}{\texttt{54,660}} & \multirow{2}{*}{\texttt{93,114}}
        & \multirow{2}{*}{\texttt{0.5217}} & \multirow{2}{*}{\texttt{3.1e-3}} & \multirow{2}{*}{\texttt{1.34}} & \multirow{2}{*}{\shortstack{\texttt{< 500B}\\ \texttt{Tokens}}} \\
        & \citep{DBLP:journals/corr/abs-2506-05209} & & & & & & & & \\

        \addlinespace

        \scalerel*{\includegraphics{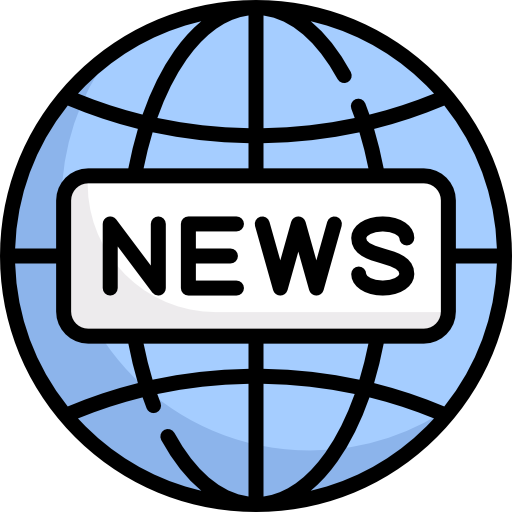}}{B}\;\texttt{Event Co-Referenced} & \texttt{MultiNews}
        & \multirow{2}{*}{\texttt{2,504}} & \multirow{2}{*}{\texttt{4,816}} & \multirow{2}{*}{\texttt{9,592}} & \multirow{2}{*}{\texttt{15,369}}
        & \multirow{2}{*}{\texttt{0.4591}} & \multirow{2}{*}{\texttt{5.7e-4}} & \multirow{2}{*}{\texttt{0.61}} & \multirow{2}{*}{\shortstack{\texttt{< 500B}\\ \texttt{Tokens}}} \\
        \quad\texttt{Documents} & \citep{DBLP:conf/acl/FabbriLSLR19} & & & & & & & & \\

        \addlinespace

        \scalerel*{\includegraphics{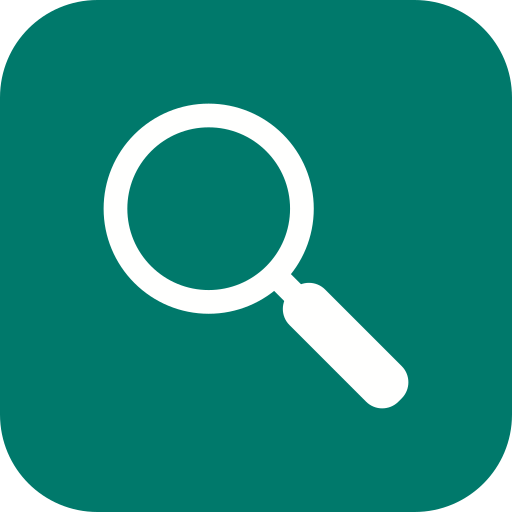}}{B}\;\texttt{Embedding Retrieval} & \texttt{NextLong}
        & \multirow{2}{*}{\texttt{147,269}} & \multirow{2}{*}{\texttt{235,444}} & \multirow{2}{*}{\texttt{282,611}} & \multirow{2}{*}{\texttt{313,076}}
        & \multirow{2}{*}{\texttt{0.5653}} & \multirow{2}{*}{\texttt{4.3e-4}} & \multirow{2}{*}{\texttt{2.24}} & \multirow{2}{*}{\shortstack{\texttt{Potentially}\\ \texttt{Unlimited}}} \\
        \quad\texttt{Augmented Documents} & \citep{DBLP:conf/icml/Gao0LZ025} & & & & & & & & \\

        \addlinespace

        \scalerel*{\includegraphics{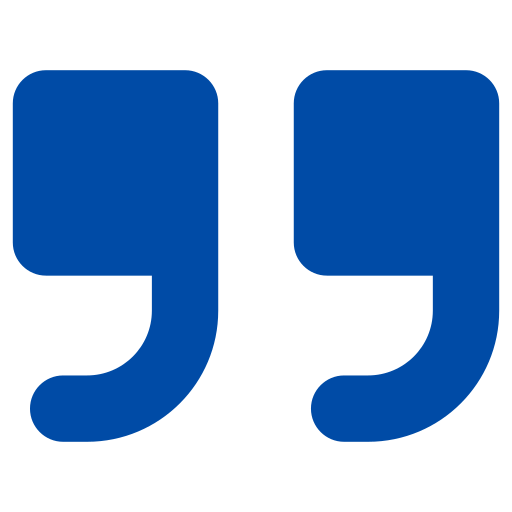}}{B}\;\texttt{Citation Retrieval} & \texttt{MegaWika 2}
        & \multirow{2}{*}{\texttt{4,012}} & \multirow{2}{*}{\texttt{7,388}} & \multirow{2}{*}{\texttt{16,656}} & \multirow{2}{*}{\texttt{24,879}}
        & \multirow{2}{*}{\texttt{0.4672}} & \multirow{2}{*}{\texttt{2.7e-3}} & \multirow{2}{*}{\texttt{0.29}} & \multirow{2}{*}{\shortstack{\texttt{< 5T}\\ \texttt{Tokens}}} \\
        \quad\texttt{Augmented Documents} & \citep{DBLP:journals/corr/abs-2508-03828} & & & & & & & & \\

        \addlinespace

        \multirow{2}{*}{\scalerel*{\includegraphics{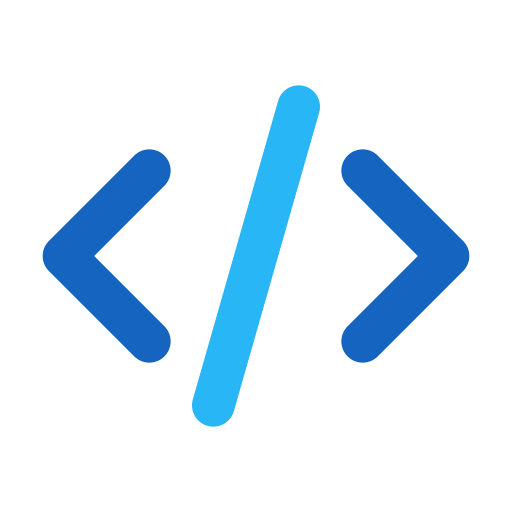}}{B}\;\texttt{In-Repo Code}} & \texttt{Stack V3}
        & \multirow{2}{*}{\texttt{4,879}} & \multirow{2}{*}{\texttt{21,868}} & \multirow{2}{*}{\texttt{49,559}} & \multirow{2}{*}{\texttt{207,681}}
        & \multirow{2}{*}{\texttt{0.5334}} & \multirow{2}{*}{\texttt{1.2e-3}} & \multirow{2}{*}{\texttt{1.82}} & \multirow{2}{*}{\shortstack{\texttt{< 10T}\\ \texttt{Tokens}}} \\
        & \citep{lozhkov2026stack-v3} & & & & & & & & \\

        \midrule

        \scalerel*{\includegraphics{Assets/Squid.png}}{B}\;\texttt{Cross-Repo Code}
        & \multirow{4}{*}{\texttt{OctoLong}} & & & & & & & & \multirow{4}{*}{\shortstack{\texttt{Potentially}\\ \texttt{Unlimited}}} \\
        \quad\texttt{-- Depth Limit 8} & & \texttt{142,046} & \texttt{298,467} & \texttt{403,450} & \texttt{487,595}
        & \texttt{0.6194} & \texttt{1.3e-4} & \texttt{6.81} & \\
        \quad\texttt{-- Depth Limit 10} & & \texttt{201,477} & \texttt{480,116} & \texttt{907,474} & \texttt{1,582,313}
        & \texttt{0.6283} & \texttt{1.2e-4} & \texttt{7.09} & \\
        \quad\texttt{-- Depth Limit 15} & & \texttt{372,559} & \texttt{808,196} & \texttt{1,747,283} & \texttt{9,779,895}
        & \texttt{0.6298} & \texttt{1.2e-4} & \texttt{7.02} & \\

        \bottomrule
    \end{tabular}
    }
    \caption{Comparison of the cross-repository contexts from the \texttt{OctoLong} pipeline (configured with varying depth limits) against existing long-context data sources. \texttt{OctoLong} flexibly sources long-context data that are (i) orders of magnitude longer and (ii) exhibit superior dependency density compared to existing long-context sources.
    }
    \label{tab:Data_Statistics}
    \vspace{-0.5em}
\end{table*}

\rparagraph{Comparison of Long Context Data Sources} Our goal is to test our hypothesis that cross-repository code contexts constitute a very good source of long and dependency-rich data. To this end, we additionally acquire preprocessed long-context corpora from a variety of traditional sources, including books, code repositories, citation graphs, and retrieved documents, and compare them against our data sourced via the \texttt{OctoLong} pipeline without token length restrictions, using BFS depth limits of 8, 10, and 15, respectively. 
A comparison of context length distributions in \cref{tab:Data_Statistics} shows that our \texttt{OctoLong} pipeline produces contexts that surpass those in any existing source by orders of magnitude. 

We also compare the density, uniformity, and salience of dependencies within the long-context data sources. This comparison is critical, as training on dependency-deprived long contexts offers limited performance gains over short-context language modeling~\citep{DBLP:conf/acl/ChenLHZSLLY24, DBLP:journals/corr/abs-2510-25804}, while substantially increasing training costs. 
We inspect all long-context corpora in \cref{tab:Data_Statistics} with a simplified LongAttn~\citep{DBLP:conf/acl/WuZZYRWSL25} score (the formula below). Specifically, we average the first attention layer scores of Phi3~\citep{DBLP:journals/corr/abs-2404-14219} and EXAONE 3.5~\citep{DBLP:journals/corr/abs-2412-04862}\footnote{For context spans longer than 128K tokens, we compute dependency saliency metrics on the trailing 128K tokens.}~\footnote{We choose two dense scorer models from model lineages outside our baseline model pool.} to calculate the mean attention mass $M$ of the last 8K query tokens against the most recent eighth, quarter, and half of the $(n-8)$K token keys from the prefix as follows (denoted with \texttt{Dep. Mean} in \cref{tab:Data_Statistics}):
\begin{equation*}
\begin{gathered}
\scalebox{0.8}{$\displaystyle
\mu_{\mathrm{Dep.}}
  = \frac{1}{24K}\sum_{j=1}^{3}\;
    \sum_{q=(n-8)K+1}^{nK}\;
    \sum_{i=b_j}^{(n-8)K} M_{i,q}
$} \\ 
\scalebox{0.7}{$\displaystyle
\text{where}\;\; b_j = (n-8)K - c_j + 1,\ \
c_{j} = \frac{(n-8)K}{2^{j}},\ \
j \in \{1,2,3\}
$}
\end{gathered}
\end{equation*}
We also track the mean attention mass fraction the final 8K tokens allot to the leading sink positions (positions 0--32; \texttt{Sink Prop.} in \cref{tab:Data_Statistics}). These metrics, in tandem, test for data-induced \textit{lost-middle} patterns~\citep{DBLP:journals/corr/abs-2510-10276, DBLP:journals/tacl/LiuLHPBPL24} where recency~\citep{Baddeley1993TheRE} and primacy~\citep{Asch1946FormingIO} biases displayed by humans spill over into the data distribution. The \texttt{LongAttn} section of \cref{tab:Data_Statistics} shows that \texttt{OctoLong} data has a markedly lower sink concentration and a notably lower incidence of lost-middle (i.e., higher \texttt{Dep. Mean}) patterns than other long-context data sources. 

Diffuse dependencies alone do not imply salience, as they may result from noisy~\citep{DBLP:conf/iclr/LeviathanKM25, DBLP:journals/corr/abs-2110-03353} or self-contradictory~\citep{DBLP:conf/acl/WuWCZZCSZL26} contexts. We thus add a simplified LongPPL~\citep{DBLP:conf/iclr/FangWLZJGD025} diagnostic, measuring the proportion of context-sensitive key tokens among the final 8K tokens: these are the tokens for which perplexity (averaged under the Phi3 and EXAONE 3.5 models) is markedly reduced when using the full context compared to just the trailing 2K tokens of the prefix (\texttt{\% Key Tokens} in \cref{tab:Data_Statistics}): 
\vspace{-0.5em}
\begin{equation*}
\begin{gathered}
\scalebox{0.8}{$\displaystyle
F_{\mathrm{key}} = \frac{1}{8K}\sum_{i=(n-8)K+1}^{nK}
\mathbf{1}\!\left[\log P_\theta(x_i \mid \boldsymbol{l}_i) - \log P_\theta(x_i \mid \boldsymbol{s}_i) > 2\right]
$} \\ 
\scalebox{0.7}{$\displaystyle
\text{where}\;\;
\boldsymbol{s}_i = \boldsymbol{x}_{(n-10)K+1:\,i-1},\ \
\boldsymbol{l}_i = \boldsymbol{x}_{1:\,i-1},\ \
i \in \{(n-8)K+1,\dots,nK\}
$}
\end{gathered}
\end{equation*}
\texttt{OctoLong}'s cross-repository code contexts in \cref{tab:Data_Statistics} contain 3--24 times more context-sensitive key tokens than other long-context corpora, including that consisting of individually serialized code repositories (\texttt{In-Repo Code}). These key tokens represent context-sensitive~\citep{DBLP:journals/corr/abs-2606-29489} and reasoning-intensive~\citep{DBLP:conf/nips/WangYGZLLDCYZLY25} points in the data that make models more amenable to post-training~\citep{Shen2026UnderstandingRF}.

\section{\texorpdfstring{\texttt{OctoLong-Instruct}}{OctoLong-Instruct} Models}
\vspace{-0.25em}
\label{sec:Setup}
We train our \texttt{OctoLong-Instruct} models in two stages: \numcircle{1} a context-extension/long-context fine-tuning (LCFT) stage, followed by \numcircle{2} a supervised fine-tuning stage (SFT), i.e., instruction tuning.

\begin{table}[t!]
    \centering
    \scalebox{0.55}{
    \begin{tabular}{lcc}
        \toprule

        \multirow{2}{*}{\scalerel*{\includegraphics{Assets/Data.png}}{B}\;\textbf{\texttt{Data Source}}} & \multicolumn{2}{c}{\textbf{\texttt{Token Prevalence (\%)}}} \\
        \cmidrule(lr){2-3}
        & \textbf{\texttt{OctoLong-LCFT}} & \textbf{\texttt{ProLong}} \\

        \midrule

        \scalerel*{\includegraphics{Assets/Code.png}}{B}\;\texttt{In-Repo \& Code-Adjacent Data} & \texttt{21.51} & \texttt{30.00} \\
        \scalerel*{\includegraphics{Assets/Paper.png}}{B}\;\texttt{Academic Papers \& PDF Data} & \texttt{18.55} & \texttt{4.00} \\
        \scalerel*{\includegraphics{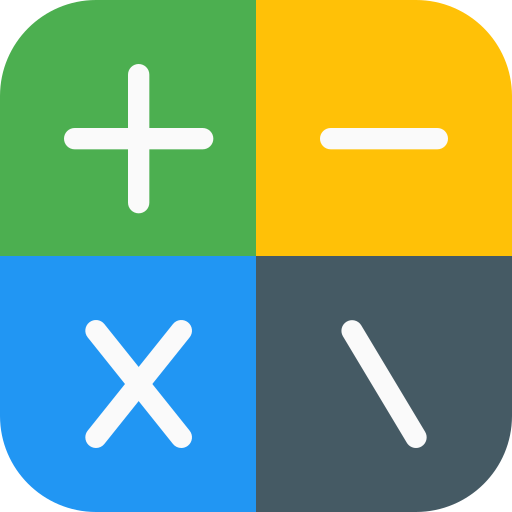}}{B}\;\texttt{Math \& Math-Adjacent Data} & \texttt{12.22} & \texttt{4.00} \\
        \scalerel*{\includegraphics{Assets/Squid.png}}{B}\;\texttt{OctoLong Data} & \texttt{12.05} & \texttt{-} \\
        \scalerel*{\includegraphics{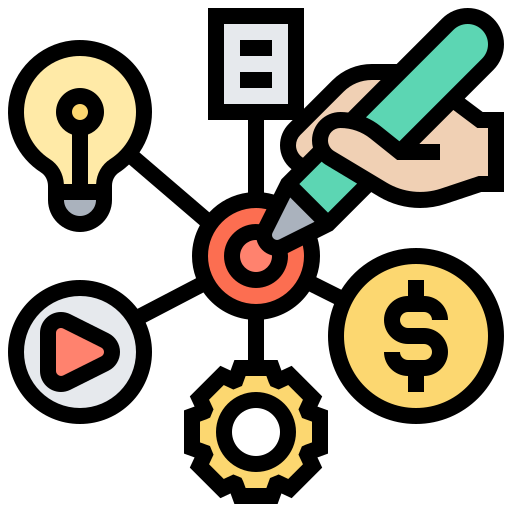}}{B}\;\texttt{General Domain Data} & \texttt{11.95} & \texttt{10.00} \\
        \scalerel*{\includegraphics{Assets/Book.png}}{B}\;\texttt{Books Data} & \texttt{9.52} & \texttt{30.00} \\
        \scalerel*{\includegraphics{Assets/Agentic.png}}{B}\;\texttt{Agentic \& Instruct-Like Data} & \texttt{6.84} & \texttt{4.00} \\
        \scalerel*{\includegraphics{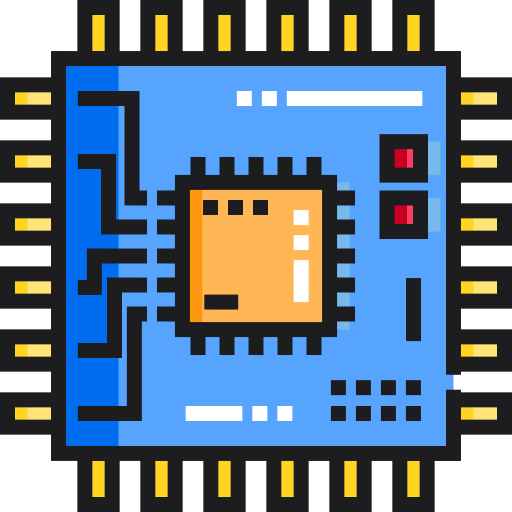}}{B}\;\texttt{Tech Content \& Forum Data} & \texttt{4.31} & \texttt{4.00} \\
        \scalerel*{\includegraphics{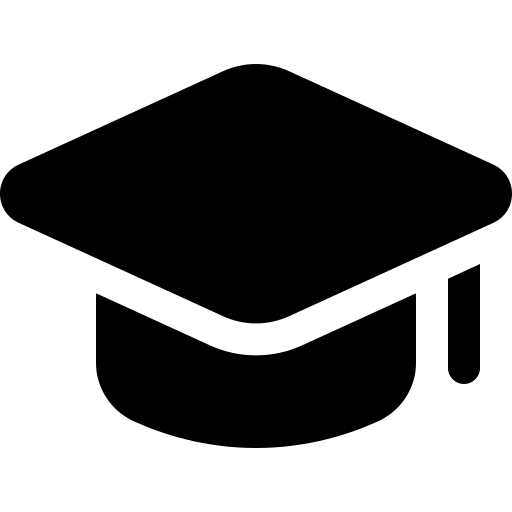}}{B}\;\texttt{Educational Data} & \texttt{3.05} & \texttt{14.00} \\

        \bottomrule
    \end{tabular}
    }
    \caption{Token prevalence by source category across the $\approx$50B-token long-context fine-tuning corpora, comparing the \texttt{OctoLong-LCFT} data mix against our control replication of ProLong~\citep{DBLP:conf/acl/0001WY025}.}
    \label{tab:Token_Prevalence}
    \vspace{-1.5em}
\end{table}

\rparagraph{Mid-Training Data} In the interest of thoroughly validating our stated hypotheses from \cref{sec:Intro}, we seek to pursue a realistically large~\citep{DBLP:journals/corr/abs-2512-13961, bakouch2025smollm3} LCFT phase of $\approx$50B tokens. We acquire $\approx$6.2B tokens of \texttt{OctoLong} data as described in \cref{sec:Dataset} and mix it with a variety of conventional short- and long-context mid-training data sources. Our auxiliary data mix spans the categories of the ProLong~\citep{DBLP:conf/acl/0001WY025} corpus. We set the length limit to 128K tokens, apply an English-only filter~\citep{DBLP:conf/vardial/FedorovaABHORS26}, and perform MinHash de-duplication~\citep{DBLP:conf/sequences/Broder97}. \cref{tab:Token_Prevalence} compares the category composition of our LCFT corpus with a size-matched, proportionally faithful replication of the ProLong corpus.

\rparagraph{SFT Data} We compile a $\approx$10B-token SFT mix, with which we fine-tune all our models (incl. the ablation variants). Although some long-context benchmarks support base model evaluations~\citep{DBLP:journals/corr/abs-2404-06654, DBLP:conf/iclr/Yen0HDFIW025, DBLP:conf/nips/KuratovBARSS024}, we choose to carry out SFT for practical reasons: most user- and industry-relevant long-context benchmarks demand instruction following~\citep{DBLP:conf/acl/WuWLSYLZ025, DBLP:conf/iclr/WuWYZCY25}, structured outputs~\citep{DBLP:journals/corr/abs-2602-03587, DBLP:conf/acl/AnG0ZLZKQ24}, and/or multi-turn interaction~\citep{DBLP:journals/corr/abs-2602-05892}. 
While earlier work suggested that a short-context SFT following LCFT suffices to post-train solid long-context LMs~\citep{DBLP:conf/acl/0001WY025, DBLP:journals/corr/abs-2406-00605}, more recent evidence points to dedicated long-sequence alignment as critical for unlocking competitive performance on generation- and reasoning-heavy long-context tasks~\citep{DBLP:journals/corr/abs-2505-17134, DBLP:conf/emnlp/BaiLZHQH0DL24, DBLP:journals/corr/abs-2505-17667, DBLP:journals/corr/abs-2502-16684, DBLP:conf/acl/YangLXJML0G25}. 
Thus, our SFT mix covers a wide distribution of domains, context lengths, and interaction turns and undergoes the same filtering pipeline as our LCFT data. We provide detailed descriptions and provenance of datasets in our final LCFT and SFT mixes in \cref{sec:appendix_data_comp}. 

\begin{figure}[t!]
    \centering
    \includegraphics[width=\linewidth, height=2.5cm]{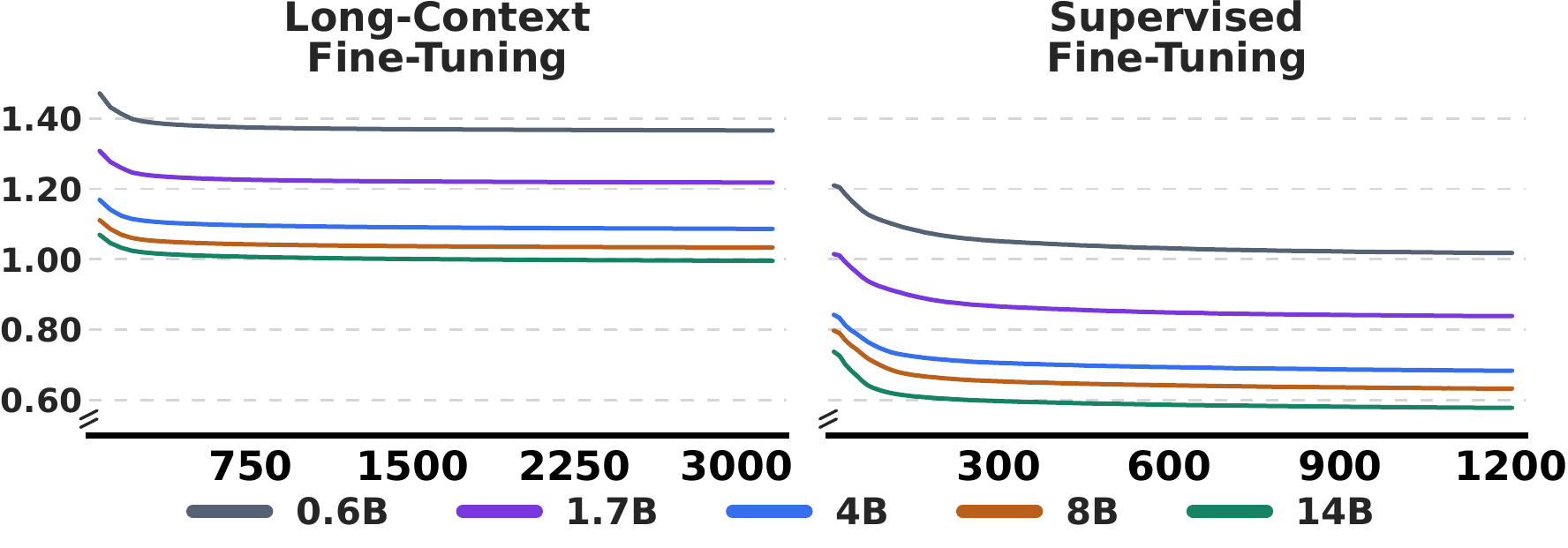}
    \caption{Scaling trends of the training loss curves (20-step moving average) across the LCFT and SFT phases.}
    \label{fig:LossPlot}
    \vspace{-1.25em}
\end{figure}

\rparagraph{Training Configuration} All \texttt{OctoLong} models descend from Qwen3~\citep{DBLP:journals/corr/abs-2505-09388} dense base models ranging from 600M to 14B parameters in size. Our LCFT phase adopts ABF RoPE scaling~\citep{DBLP:conf/naacl/XiongLMZBHMRSOK24}, raising the base frequency from the Qwen3 default of 1M to 10M while expanding the context length from 32K to 128K tokens. We avoid a multi-stage context-extension setup~\citep{DBLP:journals/corr/abs-2407-21783, DBLP:conf/nips/ZhuLWCGPK25} because our exploratory runs showed that a single-stage setup incurs no performance penalty. Although we attempt to mitigate performance regression on short inputs by mixing short- and long-context data in our mid-training mix~\citep{DBLP:conf/emnlp/ZhengLYC25}, short-input regressions can still persist~\citep{DBLP:conf/acl/DongLJXZWC25}. Hence, we follow industry practice and model-merge the pre- and post-LCFT checkpoints~\citep{arcee2025afm, bakouch2025smollm3}\footnote{We perform a 1:9 linear merge between the respective Qwen3 base and our post-LCFT checkpoints.}, obtaining the \texttt{OctoLong-Base} suite. We then perform SFT with the modified RoPE frequency and context length to obtain the \texttt{OctoLong-Instruct} suite. \Cref{fig:LossPlot} shows the loss curves of the two training stages.

All our training runs use a modified version of the \texttt{Swift}~\citep{DBLP:conf/aaai/ZhaoHHWMZJWAWZC25} framework, leveraging the \texttt{DeepSpeed-Ulysses}~\citep{DBLP:journals/corr/abs-2309-14509} sequence parallelism and \texttt{Liger} kernel optimizations~\citep{DBLP:journals/corr/abs-2410-10989} to facilitate efficient long sequence training. We use the best-fit-decreasing~\citep{DBLP:journals/jal/Baker85} scheme to pack training samples into 128K contexts for maximal efficiency.\footnote{While our choice of packing algorithm is greedy, empirically we observe a packing efficiency over 99.87\%.} We employ the AdamW~\citep{DBLP:conf/iclr/LoshchilovH19} optimizer with a peak learning rate of 1e-5 along with the WSD scheduler~\citep{DBLP:conf/nips/HageleBKAWJ24} for resumption friendliness and adopt a 5\% warmup to ensure smooth continual training~\citep{DBLP:journals/corr/abs-2308-04014}. \cref{tab:Model_Training} details all training parameters of our LCFT and SFT training stages.

\section{Results and Discussion}
\label{sec:Results}
\vspace{-0.25em}
We compare our \texttt{OctoLong-Instruct} suite of models against 18 open-weight baselines that support contexts of at least 64K tokens, across a range of long-context capability evaluations and short-context regression tests. We additionally perform three key ablations: \numcircle{1} by replacing the entire LCFT mixture with the size-matched, proportionally faithful replication of the ProLong corpus from \cref{tab:Token_Prevalence} (termed \textbf{\texttt{-Cross-Repo Code}}), \numcircle{2} skipping model-merging before the SFT stage (termed \textbf{\texttt{-Model Merging}}), and \numcircle{3} skipping LCFT and directly performing fine-tuning on our SFT mix (termed \textbf{\texttt{-Context Extension}}). 
We skip benchmark subsets that average more than 128K tokens to ensure faithful findings. All reported scores are averaged over three independent inference runs, each decoding with nucleus sampling ($p=0.95$, temperature $0.6$, and a 2K-token budget).

\begin{table*}[t!]
\centering
\scalebox{0.8}{\footnotesize \gcode\;\textbf{\texttt{Code-Specialised}} \qquad \gcext\;\textbf{\texttt{Context-Extended-After-Post-Training}} \qquad \gpost\;\textbf{\texttt{Preference/RL-Tuned}}}
\scalebox{0.55}{%
\setlength{\tabcolsep}{4pt}%
\begin{tabular}{@{}c l l r r rrr rrr@{}}
\toprule
& \multirow{3}{*}{\textbf{\texttt{Model}}} & & \multirow{3}{*}{\textbf{\texttt{Size}}} & \multicolumn{1}{c}{\multirow{3}{*}{\shortstack{\textbf{\texttt{Max}}\\ \textbf{\texttt{Length}}}}} & \multicolumn{3}{c}{\textbf{\texttt{Coding Long-Context (RQ1)}}} & \multicolumn{3}{c}{\textbf{\texttt{General Long-Context (RQ2)}}} \\
\cmidrule(lr){6-8} \cmidrule(lr){9-11}
& & & & & \hc{\textbf{\texttt{LooGLE}}} & \hc{\textbf{\texttt{LongCode}}} & \hc{\textbf{\texttt{RepoQA}}} & \hc{\textbf{\texttt{LongBench}}} & \hc{\textbf{\texttt{OpenAI}}} & \hc{\textbf{\texttt{AA-LCR}}} \\
& & & & & \hc{\textbf{\texttt{V2 Code}}} & \hc{\textbf{\texttt{QA}}} & & \hc{\textbf{\texttt{V2}}} & \hc{\textbf{\texttt{MRCR}}} & \\
\midrule
\multirow{1}{*}{\rotatebox[origin=c]{90}{\textbf{\texttt{XS}}}}
& \textbf{\hsq{\texttt{OctoLong-0.6B-Instruct}}} & \gcode & \texttt{0.6B} & \texttt{128K} & \olsc{6.87}\pmstd{0.02} & \olsc{19.31}\pmstd{0.05} & \olsc{6.12}\pmstd{0.05} & \olsc{22.82}\pmstd{0.08} & \olsc{9.21}\pmstd{0.06} & \ollcr{5.00}\pmstd{0.82} \\
\midrule
\multirow{4}{*}{\rotatebox[origin=c]{90}{\textbf{\texttt{S}}}}
& \rmodel{meta-llama/Llama-3.2-1B-Instruct} & \gpost & \texttt{1B} & \texttt{128K} & \olsc{4.83}\pmstd{0.05} & \olsc{15.90}\pmstd{0.01} & \olsc{0.65}\pmstd{0.01} & \olsc{21.11}\pmstd{0.05} & \olsc{2.53}\pmstd{0.05} & \ollcr{7.00}\pmstd{1.25} \\
& \rmodel{01-ai/Yi-Coder-1.5B-Chat} & \gcode\;\gpost & \texttt{1.5B} & \texttt{128K} & \olsc{3.32}\pmstd{0.02} & \olsc{2.41}\pmstd{0.01} & \olsc{0.81}\pmstd{0.05} & \olsc{16.48}\pmstd{0.07} & \olsc{0.59}\pmstd{0.04} & \ollcr{1.00}\pmstd{0.82} \\
& \textbf{\hsq{\texttt{OctoLong-1.7B-Instruct}}} & \gcode & \texttt{1.7B} & \texttt{128K} & \olsc{15.10}\pmstd{0.02} & \olsc{44.81}\pmstd{0.09} & \olsc{25.82}\pmstd{0.06} & \olsc{26.49}\pmstd{0.11} & \olsc{19.53}\pmstd{0.09} & \ollcr{8.00}\pmstd{0.48} \\
& \rmodel{ibm-granite/granite-3.1-2b-instruct} & \gpost & \texttt{2B} & \texttt{128K} & \olsc{15.12}\pmstd{0.08} & \olsc{36.15}\pmstd{0.21} & \olsc{24.91}\pmstd{0.13} & \olsc{24.42}\pmstd{0.14} & \olsc{13.60}\pmstd{0.14} & \ollcr{6.00}\pmstd{0.82} \\
\midrule
\multirow{4}{*}{\rotatebox[origin=c]{90}{\textbf{\texttt{M}}}}
& \rmodel{meta-llama/Llama-3.2-3B-Instruct} & \gpost & \texttt{3B} & \texttt{128K} & \olsc{23.90}\pmstd{0.12} & \olsc{18.79}\pmstd{0.06} & \olsc{6.41}\pmstd{0.05} & \olsc{22.52}\pmstd{0.08} & \olsc{15.52}\pmstd{0.12} & \ollcr{7.00}\pmstd{1.25} \\
& \rmodel{Qwen/Qwen3-4B-Instruct-2507} & \gpost & \texttt{4B} & \texttt{256K} & \olsc{28.46}\pmstd{0.08} & \olsc{63.64}\pmstd{0.01} & \olsc{57.40}\pmstd{0.09} & \olsc{29.70}\pmstd{0.09} & \olsc{38.58}\pmstd{0.11} & \ollcr{10.00}\pmstd{0.47} \\
& \textbf{\hsq{\texttt{OctoLong-4B-Instruct}}} & \gcode & \texttt{4B} & \texttt{128K} & \olsc{29.81}\pmstd{0.11} & \olsc{64.09}\pmstd{0.10} & \olsc{58.07}\pmstd{0.08} & \olsc{28.20}\pmstd{0.12} & \olsc{36.91}\pmstd{0.10} & \ollcr{10.00}\pmstd{0.82} \\
& \rmodel{arcee-ai/AFM-4.5B} & \gpost & \texttt{4.5B} & \texttt{64K} & \olsc{16.13}\pmstd{0.05} & \olsc{61.02}\pmstd{0.02} & \olsc{6.96}\pmstd{0.02} & \olsc{22.52}\pmstd{0.14} & \olsc{5.91}\pmstd{0.17} & \ollcr{7.00}\pmstd{1.63} \\
\midrule
\multirow{12}{*}{\rotatebox[origin=c]{90}{\textbf{\texttt{L}}}}
& \rmodel{Qwen/Qwen2.5-7B-Instruct-1M} & \gpost & \texttt{7B} & \texttt{1024K} & \olsc{26.91}\pmstd{0.06} & \olsc{60.96}\pmstd{0.01} & \olsc{43.45}\pmstd{0.08} & \olsc{28.24}\pmstd{0.07} & \olsc{35.30}\pmstd{0.12} & \ollcr{13.00}\pmstd{1.25} \\
& \rmodel{aws-prototyping/MegaBeam-Mistral-7B-512k} & \gcext & \texttt{7B} & \texttt{512K} & \olsc{18.16}\pmstd{0.05} & \olsc{40.22}\pmstd{0.12} & \olsc{26.22}\pmstd{0.14} & \olsc{23.56}\pmstd{0.06} & \olsc{0.00}\pmstd{0.00} & \ollcr{6.00}\pmstd{0.48} \\
& \rmodel{internlm/internlm2_5-7b-chat-1m} & \gpost & \texttt{7B} & \texttt{1024K} & \olsc{13.40}\pmstd{0.15} & \olsc{39.76}\pmstd{0.08} & \olsc{37.74}\pmstd{0.02} & \olsc{25.30}\pmstd{0.09} & \olsc{27.52}\pmstd{0.11} & \ollcr{6.00}\pmstd{1.25} \\
& \textbf{\hsq{\texttt{OctoLong-8B-Instruct}}} & \gcode & \texttt{8B} & \texttt{128K} & \olsc{29.89}\pmstd{0.14} & \olsc{65.19}\pmstd{0.44} & \olsc{64.77}\pmstd{0.22} & \olsc{39.18}\pmstd{0.16} & \olsc{40.87}\pmstd{0.19} & \ollcr{13.00}\pmstd{1.63} \\
& \rmodel{meta-llama/Llama-3.1-8B-Instruct} & \gpost & \texttt{8B} & \texttt{128K} & \olsc{23.40}\pmstd{0.11} & \olsc{51.82}\pmstd{0.05} & \olsc{34.84}\pmstd{0.08} & \olsc{28.16}\pmstd{0.12} & \olsc{30.62}\pmstd{0.07} & \ollcr{13.00}\pmstd{0.82} \\
& \rmodel{nvidia/Llama-3.1-Nemotron-8B-UltraLong-1M-Instruct} & \gcext & \texttt{8B} & \texttt{1024K} & \olsc{21.36}\pmstd{0.07} & \olsc{35.42}\pmstd{0.15} & \olsc{27.05}\pmstd{0.09} & \olsc{24.20}\pmstd{0.06} & \olsc{4.17}\pmstd{0.08} & \ollcr{10.00}\pmstd{1.70} \\
& \rmodel{princeton-nlp/Llama-3-8B-ProLong-512k-Instruct} & \gcext & \texttt{8B} & \texttt{512K} & \olsc{18.77}\pmstd{0.18} & \olsc{25.86}\pmstd{0.13} & \olsc{35.59}\pmstd{0.08} & \olsc{24.72}\pmstd{0.07} & \olsc{41.70}\pmstd{0.05} & \ollcr{6.00}\pmstd{0.48} \\
& \rmodel{ibm-granite/granite-3.1-8b-instruct} & \gpost & \texttt{8B} & \texttt{128K} & \olsc{22.23}\pmstd{0.06} & \olsc{31.57}\pmstd{0.17} & \olsc{37.99}\pmstd{0.07} & \olsc{22.49}\pmstd{0.18} & \olsc{28.31}\pmstd{0.12} & \ollcr{7.00}\pmstd{0.82} \\
& \rmodel{gradientai/Llama-3-8B-Instruct-262k} & \gcext & \texttt{8B} & \texttt{256K} & \olsc{10.96}\pmstd{0.04} & \olsc{7.12}\pmstd{0.07} & \olsc{30.85}\pmstd{0.05} & \olsc{33.47}\pmstd{0.17} & \olsc{32.83}\pmstd{0.13} & \ollcr{2.00}\pmstd{0.00} \\
& \rmodel{mistralai/Ministral-8B-Instruct-2410} & \gpost & \texttt{8B} & \texttt{128K} & \olsc{18.39}\pmstd{0.05} & \olsc{51.97}\pmstd{0.09} & \olsc{21.83}\pmstd{0.08} & \olsc{25.55}\pmstd{0.12} & \olsc{13.15}\pmstd{0.15} & \ollcr{4.00}\pmstd{0.48} \\
& \rmodel{zai-org/glm-4-9b-chat-1m} & \gpost & \texttt{9B} & \texttt{1024K} & \olsc{22.59}\pmstd{0.16} & \olsc{61.98}\pmstd{0.07} & \olsc{59.14}\pmstd{0.13} & \olsc{28.70}\pmstd{0.14} & \olsc{2.15}\pmstd{0.11} & \ollcr{12.00}\pmstd{1.25} \\
& \rmodel{01-ai/Yi-Coder-9B-Chat} & \gcode\;\gpost & \texttt{9B} & \texttt{128K} & \olsc{17.11}\pmstd{0.19} & \olsc{6.68}\pmstd{0.05} & \olsc{12.81}\pmstd{0.06} & \olsc{21.36}\pmstd{0.20} & \olsc{1.89}\pmstd{0.18} & \ollcr{5.00}\pmstd{1.25} \\
\midrule
\multirow{2}{*}{\rotatebox[origin=c]{90}{\textbf{\texttt{XL}}}}
& \textbf{\hsq{\texttt{OctoLong-14B-Instruct}}} & \gcode & \texttt{14B} & \texttt{128K} & \olsc{39.87}\pmstd{0.18} & \olsc{78.02}\pmstd{0.12} & \olsc{76.01}\pmstd{0.10} & \olsc{41.52}\pmstd{0.21} & \olsc{44.87}\pmstd{0.23} & \ollcr{21.00}\pmstd{1.70} \\
& \rmodel{Qwen/Qwen2.5-14B-Instruct-1M} & \gpost & \texttt{14B} & \texttt{1024K} & \olsc{37.17}\pmstd{0.11} & \olsc{72.10}\pmstd{0.06} & \olsc{75.06}\pmstd{0.06} & \olsc{31.44}\pmstd{0.18} & \olsc{36.62}\pmstd{0.16} & \ollcr{23.00}\pmstd{0.48} \\
\midrule
\multirow{4}{*}{\rotatebox[origin=c]{90}{\textbf{\texttt{Abl.}}}}
& \multirow{2}{*}{\textbf{\hsq{\texttt{OctoLong-8B-Instruct - Cross-Repo Code}}}} & \multirow{2}{*}{\gcode} & \multirow{2}{*}{\texttt{8B}} & \multirow{2}{*}{\texttt{128K}} & \hc{\texttt{26.38}} & \hc{\texttt{62.17}} & \hc{\texttt{61.58}} & \hc{\texttt{37.02}} & \hc{\texttt{36.88}} & \hc{\texttt{11.00}} \\
\addlinespace[-0.2em]
& & & & & \hc{\diffdo{3.51}} & \hc{\diffdo{3.02}} & \hc{\diffdo{3.19}} & \hc{\diffdo{2.16}} & \hc{\diffdo{3.99}} & \hc{\diffdo{2.00}} \\
\addlinespace[0.12em]
& \multirow{2}{*}{\textbf{\hsq{\texttt{OctoLong-8B-Instruct - Model Merging}}}} & \multirow{2}{*}{\gcode} & \multirow{2}{*}{\texttt{8B}} & \multirow{2}{*}{\texttt{128K}} & \hc{\texttt{28.12}} & \hc{\texttt{64.25}} & \hc{\texttt{63.73}} & \hc{\texttt{38.61}} & \hc{\texttt{39.20}} & \hc{\texttt{12.00}} \\
\addlinespace[-0.2em]
& & & & & \hc{\diffdo{1.77}} & \hc{\diffdo{0.94}} & \hc{\diffdo{1.04}} & \hc{\diffdo{0.57}} & \hc{\diffdo{1.67}} & \hc{\diffdo{1.00}} \\
\addlinespace[0.12em]
\bottomrule
\end{tabular}%
}
\caption{Comparison of benchmark-level averages between extant open-weight long-context LMs and the \texttt{OctoLong-Instruct} suite on our coding (\textbf{\texttt{RQ1}}: ref. \cref{subsec:RQ1}) and general-domain (\textbf{\texttt{RQ2}}: ref. \cref{subsec:RQ2}) long-context evaluation suites, followed by \texttt{OctoLong-8B-Instruct} training ablations. Per-benchmark scores are the mean over the constituent splits. Detailed per-split scores are reported in \cref{tab:Results_RQ1,tab:Results_RQ2}.}
\label{tab:Results_Main}
\vspace{-1em}
\end{table*}

\rqhead{subsec:RQ1}{Does \texttt{OctoLong} improve long-context \textit{coding} performance?} 

\vspace{0.15em}
\noindent In what constitutes the most ``intrinsic'' evaluation for \texttt{OctoLong}, we test whether training with cross-repository contexts improves models' recall and state-tracking abilities on long code inputs where relevant information occurs over a range of span sizes and dispersion settings~\citep{DBLP:conf/emnlp/GoldmanJSMDT24}.

\rrparagraph{LooGLE V2} We leverage the Code subset of LooGLE V2~\citep{DBLP:journals/corr/abs-2510-22548}. This contains the MCQ Call Graph task that tests high-dispersion call-stack dependency tracking between two call sites at the repository level, and the Version Control string similarity task that asks models to detect multi-file diffs between two repository snapshots.

\rrparagraph{LongCodeQa} We also test models' pure recall abilities in repository-level contexts using the 64K and 128K subsets of the MCQ tasks sourced from LongCodeQA~\citep {DBLP:journals/corr/abs-2505-07897}.

\rrparagraph{RepoQa} Pure-recall tests can be compromised by models using surface-level matching~\citep{DBLP:conf/icml/ModarressiDDBR025}. Hence, we audit models' ability to recall needle function implementations from repository-level contexts based on natural language description queries by evaluating them on the 64K and 128K subsets of RepoQA~\citep{DBLP:journals/corr/abs-2406-06025}. We use a BLEU score of 0.8 on recalled implementations as our acceptance threshold.

\rparagraph{Results} \cref{tab:Results_Main} displays consistently strong performance of the \texttt{OctoLong-Instruct} suite in long-context code state-tracking (Call Graph) and retrieval (Version Control and LongCodeQA) scenarios, including ones with low lexical overlap (RepoQA). This pattern holds across size classes and sequence lengths (ref. \cref{tab:Results_RQ1}).
Comparing our 8B variants trained with (\texttt{OctoLong-8B-Instruct}) and \textit{without} \textit{OctoLong} data (\texttt{-Cross-Repo Code}), we observe a substantial performance contribution of the dependency-rich cross-repository code data. We observe especially outsized gains from the \texttt{OctoLong} cross-repository contexts with high dependency density and range on the Call Graph task (ref. \cref{tab:Results_RQ1}). Poor baseline scores on this task support prior findings that high-dispersion long contexts constitute particularly challenging scenarios for LMs~\citep{DBLP:conf/acl/TianLFDLQWCZWLW25}.  
Our models show minimal performance regression between 64K and 128K settings (ref. \cref{tab:Results_RQ1}), which is rare for models at the edge of their supported context capabilities~\citep{DBLP:conf/iclr/An00LGLXK25}.  
Finally, we see signs that \texttt{OctoLong} data's performance-enhancing value extends beyond its domain. Specifically, despite restricting our data collection to Python contexts, we observe strong averages on the RepoQA task, which covers five programming languages. Moreover, this Python-only scope is incidental rather than inherent: the \texttt{OctoLong} pipeline ports to any language with a mature language server and package manager (e.g., TypeScript or Rust), making broader multilingual coverage a matter of data collection rather than a limitation of the method. 

\rqhead{subsec:RQ2}{Does \texttt{OctoLong} enhance long-context performance in the general domain?}

\vspace{0.15em}
\noindent To test the transfer effects of our \texttt{OctoLong} recipe beyond code, we run a variety of benchmarks that test in-context learning, retrieval, state tracking, and synthesis in the ``general domain''.

\rrparagraph{LongBench V2} We evaluate focused and dispersed general-domain retrieval using the single-document- and multi-document-QA MCQ subsets of LongBench V2~\citep{DBLP:conf/acl/BaiTZ0WLCX0D0L25}. We also test the Long ICL MCQ subset's performance as a measure of in-context learning on long inputs. Additionally, we stress-test the models' conversational state-tracking capabilities using the Long Dialog MCQ subset. We benchmark on the 32K--128K (medium) length portions of all subsets.

\rrparagraph{OpenAi Mrcr} We leverage the 64K and 128K capped-length subsets of the 2-Needle multi-round co-reference resolution task published by OpenAI MRCR~\citep{openai2025mrcr}. This string similarity task requires models to recall the hashed answer contents for a query about a multi-turn chat. It realistically evaluates models' recall in high-dispersion settings with multiple distractors.

\rrparagraph{Aa-Lcr} Long-context evaluations largely center around direct and indirect retrieval, which, however challenging, can hide large regressions in models' reasoning abilities in free-form generation~\citep{DBLP:journals/corr/abs-2501-15089, DBLP:journals/corr/abs-2604-07981}. To address this possibility, we evaluate on AA-LCR~\citep{artificialanalysis2025lcr}, an industry-standard test that requires models to perform multi-hop reasoning and inference in-context on documents from a variety of domains. Scoring relies on \texttt{pass@1} answer matching using a \texttt{Qwen3-235B-A22B-Instruct-2507} judge model.

\rparagraph{Results} \cref{tab:Results_Main} summarizes the results that answer our RQ2. Our LCFT data mixture brings strong results across the board on general-domain long-context tasks. In particular, the \texttt{OctoLong-Instruct} suite compares favorably against the \texttt{-Cross-Repo Code} ablation, highlighting how training on cross-repository contexts transfers positively to the general domain. Our \texttt{OctoLong-Instruct-4B} fares competitively against \texttt{Qwen3-4B-Instruct-2507}, a strong context-extended and RL-post-trained Qwen3 model, while dominating all other size classes. Encouragingly, our recipe performs strongly on the industry-relevant MRCR and AA-LCR benchmarks, suggesting that the dependency-rich data we train on yields tangible gains in the most challenging, distractor-laden, high-dispersion, and reasoning-intensive settings. 
We also note the non-negligible contribution of model merging (\texttt{-Model Merging} performance drops) in the general domain.


\begin{table*}[t!]
\centering
\scalebox{0.8}{\footnotesize \gcode\;\textbf{\texttt{Code-Specialised}} \qquad \gcext\;\textbf{\texttt{Context-Extended-After-Post-Training}} \qquad \gpost\;\textbf{\texttt{Preference/RL-Tuned}}}
\scalebox{0.55}{%
\setlength{\tabcolsep}{4pt}%
\begin{tabular}{@{}c l l r r rr rr@{}}
\toprule
& \multirow{3}{*}{\textbf{\texttt{Model}}} & & \multirow{3}{*}{\textbf{\texttt{Size}}} & \multicolumn{1}{c}{\multirow{3}{*}{\shortstack{\textbf{\texttt{Max}}\\ \textbf{\texttt{Length}}}}} & \multicolumn{2}{c}{\textbf{\texttt{Agentic Tool Use (RQ3)}}} & \multicolumn{2}{c}{\textbf{\texttt{Short Coding (RQ4)}}} \\
\cmidrule(lr){6-7} \cmidrule(lr){8-9}
& & & & & \hc{\textbf{\texttt{BFCL}}} & \hc{\textbf{\texttt{$\tau^2$}}} & \hc{\textbf{\texttt{LiveCode}}} & \hc{\textbf{\texttt{BCB Hard}}} \\
& & & & & \hc{\textbf{\texttt{V3}}} & \hc{\textbf{\texttt{Bench}}} & \hc{\textbf{\texttt{Bench V6}}} & \hc{\textbf{\texttt{Instruct}}} \\
\midrule
\multirow{1}{*}{\rotatebox[origin=c]{90}{\textbf{\texttt{XS}}}}
& \textbf{\hsq{\texttt{OctoLong-0.6B-Instruct}}} & \gcode & \texttt{0.6B} & \texttt{128K} & \olsc{40.46}\pmstd{0.26} & \olsc{15.89}\pmstd{0.21} & \olcg{4.55}\pmstd{0.04} & \olcg{1.60}\pmstd{0.09} \\
\midrule
\multirow{4}{*}{\rotatebox[origin=c]{90}{\textbf{\texttt{S}}}}
& \rmodel{meta-llama/Llama-3.2-1B-Instruct} & \gpost & \texttt{1B} & \texttt{128K} & \olsc{19.72}\pmstd{0.15} & \olsc{13.64}\pmstd{0.24} & \olcg{2.45}\pmstd{0.06} & \olcg{0.80}\pmstd{0.07} \\
& \rmodel{01-ai/Yi-Coder-1.5B-Chat} & \gcode\;\gpost & \texttt{1.5B} & \texttt{128K} & \olsc{14.89}\pmstd{0.22} & \olsc{17.34}\pmstd{0.29} & \olcg{8.91}\pmstd{0.05} & \olcg{5.11}\pmstd{0.08} \\
& \textbf{\hsq{\texttt{OctoLong-1.7B-Instruct}}} & \gcode & \texttt{1.7B} & \texttt{128K} & \olsc{48.25}\pmstd{0.20} & \olsc{17.01}\pmstd{0.28} & \olcg{10.79}\pmstd{0.14} & \olcg{4.90}\pmstd{0.19} \\
& \rmodel{ibm-granite/granite-3.1-2b-instruct} & \gpost & \texttt{2B} & \texttt{128K} & \olsc{41.86}\pmstd{0.16} & \olsc{17.94}\pmstd{0.23} & \olcg{9.57}\pmstd{0.11} & \olcg{6.21}\pmstd{0.14} \\
\midrule
\multirow{4}{*}{\rotatebox[origin=c]{90}{\textbf{\texttt{M}}}}
& \rmodel{meta-llama/Llama-3.2-3B-Instruct} & \gpost & \texttt{3B} & \texttt{128K} & \olsc{24.22}\pmstd{0.19} & \olsc{14.63}\pmstd{0.38} & \olcg{8.33}\pmstd{0.22} & \olcg{5.02}\pmstd{0.24} \\
& \rmodel{Qwen/Qwen3-4B-Instruct-2507} & \gpost & \texttt{4B} & \texttt{256K} & \olsc{57.81}\pmstd{0.27} & \olsc{30.85}\pmstd{0.27} & \olcg{17.86}\pmstd{0.18} & \olcg{19.87}\pmstd{0.24} \\
& \textbf{\hsq{\texttt{OctoLong-4B-Instruct}}} & \gcode & \texttt{4B} & \texttt{128K} & \olsc{58.45}\pmstd{0.36} & \olsc{25.21}\pmstd{0.42} & \olcg{17.44}\pmstd{0.26} & \olcg{16.31}\pmstd{0.29} \\
& \rmodel{arcee-ai/AFM-4.5B} & \gpost & \texttt{4.5B} & \texttt{64K} & \olsc{43.29}\pmstd{0.19} & \olsc{20.15}\pmstd{0.22} & \olcg{5.59}\pmstd{0.12} & \olcg{2.14}\pmstd{0.21} \\
\midrule
\multirow{12}{*}{\rotatebox[origin=c]{90}{\textbf{\texttt{L}}}}
& \rmodel{Qwen/Qwen2.5-7B-Instruct-1M} & \gpost & \texttt{7B} & \texttt{1024K} & \olsc{51.23}\pmstd{0.24} & \olsc{23.02}\pmstd{0.22} & \olcg{16.45}\pmstd{0.17} & \olcg{13.11}\pmstd{0.20} \\
& \rmodel{aws-prototyping/MegaBeam-Mistral-7B-512k} & \gcext & \texttt{7B} & \texttt{512K} & \olsc{21.23}\pmstd{0.29} & \olsc{0.00}\pmstd{0.00} & \olcg{2.77}\pmstd{0.42} & \olcg{0.52}\pmstd{0.23} \\
& \rmodel{internlm/internlm2_5-7b-chat-1m} & \gpost & \texttt{7B} & \texttt{1024K} & \olsc{28.23}\pmstd{0.33} & \olsc{22.38}\pmstd{0.21} & \olcg{2.82}\pmstd{0.23} & \olcg{4.23}\pmstd{0.35} \\
& \textbf{\hsq{\texttt{OctoLong-8B-Instruct}}} & \gcode & \texttt{8B} & \texttt{128K} & \olsc{61.67}\pmstd{0.54} & \olsc{27.35}\pmstd{0.41} & \olcg{20.82}\pmstd{0.31} & \olcg{19.44}\pmstd{0.39} \\
& \rmodel{meta-llama/Llama-3.1-8B-Instruct} & \gpost & \texttt{8B} & \texttt{128K} & \olsc{30.99}\pmstd{0.47} & \olsc{16.92}\pmstd{0.18} & \olcg{11.83}\pmstd{0.22} & \olcg{11.21}\pmstd{0.27} \\
& \rmodel{nvidia/Llama-3.1-Nemotron-8B-UltraLong-1M-Instruct} & \gcext & \texttt{8B} & \texttt{1024K} & \olsc{26.98}\pmstd{0.28} & \olsc{16.87}\pmstd{0.19} & \olcg{14.34}\pmstd{0.11} & \olcg{10.35}\pmstd{0.24} \\
& \rmodel{princeton-nlp/Llama-3-8B-ProLong-512k-Instruct} & \gcext & \texttt{8B} & \texttt{512K} & \olsc{39.00}\pmstd{0.31} & \olsc{21.07}\pmstd{0.38} & \olcg{5.12}\pmstd{0.23} & \olcg{7.21}\pmstd{0.39} \\
& \rmodel{ibm-granite/granite-3.1-8b-instruct} & \gpost & \texttt{8B} & \texttt{128K} & \olsc{49.11}\pmstd{0.28} & \olsc{21.09}\pmstd{0.36} & \olcg{12.87}\pmstd{0.09} & \olcg{13.12}\pmstd{0.22} \\
& \rmodel{gradientai/Llama-3-8B-Instruct-262k} & \gcext & \texttt{8B} & \texttt{256K} & \olsc{20.08}\pmstd{0.25} & \olsc{19.24}\pmstd{0.49} & \olcg{3.42}\pmstd{0.15} & \olcg{3.68}\pmstd{0.19} \\
& \rmodel{mistralai/Ministral-8B-Instruct-2410} & \gpost & \texttt{8B} & \texttt{128K} & \olsc{10.38}\pmstd{0.24} & \olsc{8.97}\pmstd{0.37} & \olcg{13.23}\pmstd{0.33} & \olcg{14.21}\pmstd{0.38} \\
& \rmodel{zai-org/glm-4-9b-chat-1m} & \gpost & \texttt{9B} & \texttt{1024K} & \olsc{33.87}\pmstd{0.39} & \olsc{21.63}\pmstd{0.28} & \olcg{13.62}\pmstd{0.29} & \olcg{11.33}\pmstd{0.41} \\
& \rmodel{01-ai/Yi-Coder-9B-Chat} & \gcode\;\gpost & \texttt{9B} & \texttt{128K} & \olsc{49.91}\pmstd{0.27} & \olsc{20.78}\pmstd{0.34} & \olcg{17.45}\pmstd{0.12} & \olcg{13.04}\pmstd{0.28} \\
\midrule
\multirow{2}{*}{\rotatebox[origin=c]{90}{\textbf{\texttt{XL}}}}
& \textbf{\hsq{\texttt{OctoLong-14B-Instruct}}} & \gcode & \texttt{14B} & \texttt{128K} & \olsc{67.38}\pmstd{0.52} & \olsc{34.70}\pmstd{0.78} & \olcg{22.43}\pmstd{0.31} & \olcg{21.52}\pmstd{0.44} \\
& \rmodel{Qwen/Qwen2.5-14B-Instruct-1M} & \gpost & \texttt{14B} & \texttt{1024K} & \olsc{55.81}\pmstd{0.35} & \olsc{26.12}\pmstd{0.45} & \olcg{20.89}\pmstd{0.23} & \olcg{19.61}\pmstd{0.29} \\
\midrule
\multirow{8}{*}{\rotatebox[origin=c]{90}{\textbf{\texttt{Abl.}}}}
& \multirow{2}{*}{\textbf{\hsq{\texttt{OctoLong-8B-Instruct - Cross-Repo Code}}}} & \multirow{2}{*}{\gcode} & \multirow{2}{*}{\texttt{8B}} & \multirow{2}{*}{\texttt{128K}} & \hc{\texttt{57.74}} & \hc{\texttt{22.38}} & \hc{\texttt{19.67}} & \hc{\texttt{17.56}} \\
\addlinespace[-0.2em]
& & & & & \hc{\diffdo{3.93}} & \hc{\diffdo{4.97}} & \hc{\diffdo{1.15}} & \hc{\diffdo{1.88}} \\
\addlinespace[0.12em]
& \multirow{2}{*}{\textbf{\hsq{\texttt{OctoLong-8B-Instruct - Model Merging}}}} & \multirow{2}{*}{\gcode} & \multirow{2}{*}{\texttt{8B}} & \multirow{2}{*}{\texttt{128K}} & \hc{\texttt{61.35}} & \hc{\texttt{26.91}} & \hc{\texttt{17.75}} & \hc{\texttt{18.04}} \\
\addlinespace[-0.2em]
& & & & & \hc{\diffdo{0.32}} & \hc{\diffdo{0.44}} & \hc{\diffdo{3.07}} & \hc{\diffdo{1.40}} \\
\addlinespace[0.12em]
& \multirow{2}{*}{\textbf{\hsq{\texttt{OctoLong-8B-Instruct - Context Extension}}}} & \multirow{2}{*}{\gcode} & \multirow{2}{*}{\texttt{8B}} & \multirow{2}{*}{\texttt{32K}} & \hc{\texttt{-}} & \hc{\texttt{-}} & \hc{\texttt{20.49}} & \hc{\texttt{19.03}} \\
\addlinespace[-0.2em]
& & & & &  &  & \hc{\diffdo{0.33}} & \hc{\diffdo{0.41}} \\
\addlinespace[0.12em]
& \multirow{2}{*}{\textbf{\hsq{\texttt{Qwen3-8B}}}} & \multirow{2}{*}{\gpost} & \multirow{2}{*}{\texttt{8B}} & \multirow{2}{*}{\texttt{32K}} & \hc{\texttt{-}} & \hc{\texttt{-}} & \hc{\texttt{23.18}} & \hc{\texttt{20.73}} \\
\addlinespace[-0.2em]
& & & & &  &  & \hc{\diffup{2.36}} & \hc{\diffup{1.29}} \\
\addlinespace[0.12em]
\bottomrule
\end{tabular}%
}
\caption{Comparison of benchmark-level summaries between extant open-weight long-context LMs and the \texttt{OctoLong-Instruct} suite on our agentic tool-use (\textbf{\texttt{RQ3}}: ref. \cref{subsec:RQ3}) and short-context coding (\textbf{\texttt{RQ4}}: ref. \cref{subsec:RQ4}) evaluation suites, followed by \texttt{OctoLong-8B-Instruct} training ablations. Agentic tool-use evaluations report the mean over their constituent settings, while short-context coding evaluations report the \texttt{Pass@1} score. Detailed per-split scores for \textbf{\texttt{RQ3}} are reported in \cref{tab:Results_RQ3} and higher pass-rate scores for \textbf{\texttt{RQ4}} are reported in \cref{tab:Results_RQ4}.}
\label{tab:Results_Agent}
\vspace{-0.7em}
\end{table*}

\rqhead{subsec:RQ3}{Do \texttt{OctoLong} gains extend downstream to agentic use-cases?}

\vspace{0.15em}
\noindent Models' capabilities in multi-turn long-horizon interactions with their environment~\citep{DBLP:conf/icml/JhaAWYCCBVKKZTP25, DBLP:journals/corr/abs-2603-08640, DBLP:journals/corr/abs-2510-04374} arguably represent the primary value proposition behind the recent trend towards increasing context lengths~\citep{DBLP:conf/nips/KwaWBDGHJKRABBD25}. Hence, we directly evaluate how our LCFT recipe performs post-SFT on downstream agentic tasks that require tool use across a range of turn budgets and difficulty levels.

\rrparagraph{Bfcl V3} We evaluate models on the live, non-live, and multi-turn splits of BFCL V3~\citep{DBLP:conf/icml/PatilMYJSSG25}. The benchmark stress-tests models in single- and multi-turn tool calling while also evaluating resilience against common pathological scenarios such as missing arguments and shifting live APIs.

\rrparagraph{$\tau^2$-Bench} We further report \texttt{pass@1} scores on $\tau^2$-Bench~\citep{DBLP:journals/corr/abs-2506-07982}, a challenging tool-calling benchmark consisting of underspecified queries that supply numerous domain-specific tools and scores based on state equivalence in an environment-level database. Along with multi-turn capability, the benchmark demands long-context multi-step support with several simulated-user turns per trajectory.\footnote{We use OpenAI gpt-5.1 as the user simulator.} Additionally, the evaluation also features dual-control scenarios where the simulated user makes concurrent changes to the shared environment state, thus constituting a realistic proxy for long-horizon user-LM cooperation.

\rparagraph{Results} \cref{tab:Results_Agent} shows the competitive to outstanding performance of the \texttt{OctoLong-Instruct} suite across size classes, often against baselines that have undergone extensive dedicated post-training for multi-turn interaction~\citep{DBLP:journals/corr/abs-2501-15383, DBLP:journals/corr/abs-2403-04652, DBLP:conf/acl/WuS25}. 
Crucially, we observe the greatest performance contribution from our cross-repository \texttt{OctoLong} data here on the agentic tool-use tasks (i.e., we observe the largest performance drops for the \texttt{-Cross-Repo Code} variant). This suggests that our context-extension recipe avoids long-form generation pathologies common for many long-context LMs~\citep{DBLP:conf/emnlp/DuTRRBGWSHP25}.

\rqhead{subsec:RQ4}{Does \texttt{OctoLong} preserve short-context coding and API-use abilities?}

\vspace{0.15em}
\noindent Finally, we use short-context, execution-validated code-generation tests to profile how training on long cross-repository data affects models' performance on short-context coding tasks. On all benchmarks, we report models' \texttt{pass@\{1,5,10,25\}} on a sampling group size $N$=50.

\rrparagraph{LiveCodeBench V6} We benchmark algorithmic coding capabilities by sourcing code judge problems from LiveCodeBench V6~\citep{DBLP:conf/iclr/JainHGLYZWSSS25}.\footnote{To minimize the probability of our auxiliary data being contaminated, we only test on the new code contest problems released in the V6 split (Dec 2024--April 2025), rather than the cumulative setup.}

\rrparagraph{BigCodeBench Hard} The \texttt{OctoLong} pipeline often sources function and class call sites with their reference implementation together in the same context. Prior work shows that this improves API usage~\citep{DBLP:journals/corr/abs-2605-17957}. We test this directly via BigCodeBench Instruct Hard~\citep{DBLP:conf/iclr/ZhuoVCH0WYZHPB025}. 

\rparagraph{Results} \cref{tab:Results_Agent} shows that \texttt{OctoLong-Instruct} models yield strong results on short-length code generation across all size classes. Our context extension largely retains short-length coding abilities (ref. \texttt{-Context Extension} ablation for the 8B model). Our models surpass several baselines post-trained on code generation and match the post-trained \texttt{Qwen3-8B} model at lower sampling budgets, while comfortably surpassing it at higher budgets (ref. \cref{tab:Results_RQ4}). 
This discrepancy likely stems from distribution sharpening during RL post-training~\citep{DBLP:conf/iclr/KirkMNLHGR24, DBLP:journals/corr/abs-2601-15609}. 
The \texttt{-Cross-Repo Code} ablation results also isolate the specific benefits of \texttt{OctoLong} data, especially for the API use tasks. The \texttt{-Model Merging} ablation renders merging with a pre-LCFT checkpoint effective for retaining short-context performance. Finally, the comparative results of Llama 3.1, ProLong~\citep{DBLP:conf/acl/0001WY025}, and UltraLong~\citep{DBLP:conf/acl/XuPXLWSLC26} show that extending context on an already post-trained model is suboptimal and can degrade short-context performance.

\section{Conclusion}
\label{sec:Conclusion}
\vspace{-0.25em}

In this work, we present \texttt{OctoLong}, an effort to engineer dependency-dense long-context corpora via recursive retrieval of code \textit{across} repository boundaries. We detail a pipeline that runs AST queries in tandem with a language server backend inside a dependency-provisioned containerized environment to source referenced class and function implementations when feasible. We use this pipeline to curate $\approx$6.2B tokens worth of high-quality dependency-rich contexts capped at 128K tokens in length. We then mix our cross-repository contexts with data from conventional long-context sources to compile the $\approx$50B-token \texttt{OctoLong-LCFT} corpus. This corpus, in conjunction with \texttt{OctoLong-SFT}, our $\approx$10B-token curated mix of agentic and instruction-tuning data, enables us to develop the \texttt{OctoLong-Instruct} suite of 128K-context models (derived from Qwen3), with between 600M and 14B parameters. Extensive experiments and ablations show that context extension with dependency-rich code data improves context retrieval, state tracking, and repository-level code understanding while unlocking superior agentic capabilities and API usage. 
We hope our work enables larger context lengths with minimal context rot~\citep{hong2025context}, empowering models in emergent fields like test-time training~\citep{DBLP:journals/corr/abs-2512-23675} and self-improvement~\citep{DBLP:conf/acl/YinWPL0W25}.

\section*{Limitations}
\vspace{-0.25em}

We show the effectiveness of the \texttt{OctoLong} pipeline by conducting long-context and supervised fine-tuning on realistically large corpora of 50B and 10B tokens, respectively. However, our study does not directly verify whether the gains of training on cross-repository contexts carry over post-RL. We deliberately make this specific compute trade-off by deferring to prior evidence outlining how the vast majority of knowledge learned and the circuits induced during preceding training phases are retained after RL~\citep{DBLP:journals/corr/abs-2509-25758, DBLP:journals/corr/abs-2509-04259, DBLP:journals/corr/abs-2511-08567}. Size-controlled comparisons of the \texttt{OctoLong-Instruct} models against several RL-trained long-context baselines in \cref{sec:Results} outline our recipe's general competitiveness. 

We also simplify our training runs by omitting low-resource language data, limiting model sizes to 14B parameters and restricting context lengths to 128K tokens. Each represents a non-trivial departure from how most frontier LMs operate today. However, important signs suggest that \texttt{OctoLong} should transfer well along these axes. First, prior methods for engineering long contexts also omit multilingual long-context data~\citep{DBLP:conf/acl/0001WY025, DBLP:conf/icml/FuPNYHK024} owing to its rarity~\citep{harm2024context} but generalize well to such settings nonetheless~\citep{DBLP:conf/naacl/HengleBDC25, DBLP:journals/corr/abs-2503-01996}. Second, our experiments in \cref{sec:Results} show how the \texttt{OctoLong-Instruct} models become increasingly adept at longer contexts as they grow, following trends described in the literature~\citep{DBLP:journals/tacl/LiuLHPBPL24, DBLP:conf/iclr/Liu0AQL24}. Third, our analyses in \cref{sec:Dataset} show how \texttt{OctoLong} can source code contexts of millions of tokens without sacrificing dependency density or salience.

Finally, we note that the recursive dependency retrieval in \texttt{OctoLong} is limited to within-language code where the language server backend can readily leverage package manager conventions to locate implementations. Our pipeline likely does not account for important dependencies in complex codebases that contain subroutines compiled in a different language that the calling code interacts with using an application binary interface (ABI) or a foreign function interface (FFI). Handling these cases requires polyglot static analysis and control-flow tracking, an active area of research~\citep{DBLP:conf/iceccs/HoudailleKBJC23, DBLP:journals/corr/abs-2108-02961}.

\makeatletter
\ifacl@finalcopy
\section*{Acknowledgments}

We gratefully acknowledge support from the hessian.AI Service Center (funded by the German Federal Ministry of Research, Technology, and Space, BMFTR, grant no. 16IS22091) and the hessian.AI Innovation Lab (funded by the Hessian Ministry for Digital Strategy and Innovation, grant no. S-DIW04/0013/003). Additionally, our work is supported by the German Federal Ministry of Education and Research and the Hessian Ministry of Higher Education, Research, Science and the Arts within their joint support of the National Research Center for Applied Cybersecurity ATHENE alongside the Federal Ministry of Research, Technology and Space within the Jupyter AI Factory. We also thank the Gauss Center for Supercomputing e.V. (GCS) for providing computing time on the Supercomputer JUWELS at the Jülich Supercomputing Center supported by GCS through funding by the German Federal Ministry of Research, Technology and Space and the Ministry of Culture and Science of the State of North Rhine-Westphalia.

\fi
\makeatother


\bibliography{custom}

\newpage
\appendix

\section{Architecture \& Training Details}
\label{sec:appendix_model_training}

\providecommand{\tbd}{{\color{DeepBlood}\texttt{?}}}
\providecommand{\hficon}{\raisebox{-1.5pt}{\includegraphics[height=9pt]{Assets/Hf.png}}}
\providecommand{\hfmodel}[1]{\hficon\;\href{https://huggingface.co/#1}{\texttt{#1}}}
\providecommand{\hfcollection}[2]{\hficon\;\href{https://huggingface.co/collections/#1}{\texttt{#2}}}

\newlength{\mtcolwd}
\settowidth{\mtcolwd}{\texttt{14B:}\;\texttt{10,100}}
\newcommand{\mtsizecol}[1]{\makebox[\mtcolwd][l]{\begin{tabular}[t]{@{}r@{\;}l@{}}#1\end{tabular}}}

\begin{table}[ht!]
    \centering
    \scalebox{0.55}{
    \begin{tabular}{lc}
        \toprule

        \textbf{\texttt{Attribute}} & \textbf{\texttt{Value}} \\

        \midrule

        \multicolumn{2}{c}{\textbf{\texttt{Common Attributes}}} \\

        \midrule

        \texttt{Backbone Family} & \hfcollection{Qwen/qwen3}{Qwen/Qwen3} \\
        \texttt{Backbone Type} & \texttt{Dense} \\
        \texttt{Backbone Sizes} & \texttt{0.6B, 1.7B, 4B, 8B, 14B} \\
        \texttt{Tuning Strategy} & \texttt{Full Fine-Tuning} \\
        \texttt{Training Sequence Length} & \texttt{131,072 Tokens} \\
        \texttt{Model Datatype} & \texttt{bfloat16} \\
        \texttt{Flash Attention Variant} & \texttt{Version 2} \\
        \texttt{FSDP Variant} & \texttt{Version 2} \\
        \texttt{Sequence Parallelism Mode} & \texttt{DeepSpeed-Ulysses} \\
        \texttt{Sequence Parallelism Degree} &
        \mtsizecol{
            \texttt{0.6B:} & \texttt{2} \\
            \texttt{1.7B:} & \texttt{2} \\
            \texttt{4B:}   & \texttt{2} \\
            \texttt{8B:}   & \texttt{4} \\
            \texttt{14B:}  & \texttt{4} \\
        } \\
        \texttt{Liger Kernels} & \texttt{True} \\
        \texttt{Gradient Checkpointing} & \texttt{True} \\
        \texttt{Optimizer Type} & \texttt{AdamW-Fused} \\
        \texttt{Beta} & \texttt{\{0.9, 0.95\}} \\
        \texttt{Epsilon} & \texttt{1e-8} \\
        \texttt{Weight Decay} & \texttt{0.1} \\
        \texttt{Gradient Clipping} & \texttt{1.0} \\
        \texttt{Scheduler Type} & \texttt{Warmup-Stable-Decay} \\
        \texttt{Scheduler Warmup Proportion} & \texttt{0.05} \\
        \texttt{Peak Learning Rate} & \texttt{1e-5} \\
        \texttt{Terminal Learning Rate Ratio} & \texttt{0.02} \\
        \texttt{Per-Device Batch Size} & \texttt{1} \\
        \texttt{Training Epochs} & \texttt{1.0} \\
        \texttt{Random Seed} & \texttt{77} \\

        \midrule

        \multicolumn{2}{c}{\textbf{\texttt{LCFT-Specific Attributes (OctoLong-Base)}}} \\

        \midrule

        \texttt{Training Corpus} & \texttt{OctoLong-LCFT} \\
        \texttt{Training Tokens} & \texttt{51.5B} \\
        \texttt{Optimization Steps} & \texttt{3072} \\
        \texttt{Scheduler Decay Steps} & \texttt{0} \\
        \texttt{Global Batch Size} & \texttt{16.8M Tokens} \\
        \texttt{Training GPU Count} &
        \mtsizecol{
            \texttt{0.6B:} & \texttt{32} \\
            \texttt{1.7B:} & \texttt{64} \\
            \texttt{4B:}   & \texttt{64} \\
            \texttt{8B:}   & \texttt{128} \\
            \texttt{14B:}  & \texttt{128} \\
        } \\

        \midrule

        \multicolumn{2}{c}{\textbf{\texttt{SFT-Specific Attributes (OctoLong-Instruct)}}} \\

        \midrule

        \texttt{Initialization} & \texttt{OctoLong-*-Base} \\
        \texttt{Training Corpus} & \texttt{OctoLong-SFT} \\
        \texttt{Training Tokens} & \texttt{10.1B} \\
        \texttt{Optimization Steps} & \texttt{1202} \\
        \texttt{Scheduler Decay Steps} & \texttt{75} \\
        \texttt{Global Batch Size} & \texttt{8.4M Tokens} \\
        \texttt{Training GPU Count} &
        \mtsizecol{
            \texttt{0.6B:} & \texttt{16} \\
            \texttt{1.7B:} & \texttt{32} \\
            \texttt{4B:}   & \texttt{64} \\
            \texttt{8B:}   & \texttt{64} \\
            \texttt{14B:}  & \texttt{128} \\
        } \\

        \bottomrule
    \end{tabular}
    }
    \caption{Architectural heritage and training attributes of the \texttt{OctoLong} suite of models across the LCFT and SFT stages. We outline the training data curation in \cref{sec:Setup} and detail its composition in \cref{sec:appendix_data_comp}.}
    \label{tab:Model_Training}
\end{table}

\section{Data Composition \& Processing}
\label{sec:appendix_data_comp}

All constituent datasets in both our LCFT and SFT mixtures are subjected to a common cleaning and decontamination pipeline. Specifically, we first apply an English-only language filter using the OpenLID~\citep{DBLP:conf/vardial/FedorovaABHORS26} classifier with a confidence threshold of 0.9, then apply a MinHash de-duplication~\citep{DBLP:conf/sequences/Broder97} filter with a shingle size of 20 and a Jaccard similarity threshold of 0.7. Next, following prior work~\citep{DBLP:conf/nips/BrownMRSKDNSSAA20, DBLP:conf/iclr/ElazarBMRSSWGS024}, we decontaminate against all of our downstream evaluations by discarding any sample whose text registers a \texttt{13-gram} overlap with a test-set instance. We serialize every function-call interaction in the SFT data into the Hermes~\citep{DBLP:journals/corr/abs-2408-11857} format to ensure smooth operation with the Qwen3 chat templates. Additionally, for any dataset that ships intermediate chain-of-thought or reasoning traces, we retain only the final response and discard the thinking blocks.

\subsection{LCFT Data Mixture}
\label{subsec:lcft_data_comp}

Our LCFT corpus comprises $\approx$50B tokens spanning the nine source categories outlined in \cref{tab:Token_Prevalence}. We detail the data sources that feed the category-level document pools used to create \texttt{OctoLong-LCFT} and our replication of the ProLong~\citep{DBLP:conf/acl/0001WY025} corpus. We also state the sample counts that each source contributes to the final \texttt{OctoLong-LCFT} dataset.

\dscat{Assets/Squid.png}{OctoLong Data}

\begin{enumerate}[label=\protect\numcircle{\arabic*}, leftmargin=*]

\dataset
  {OctoLong}
  {}
  {57993}
  {Our own cross-repository code contexts, averaging $\approx$107K tokens in length, mined and assembled as described in \cref{sec:Dataset}, containing $\approx$6.2B tokens of dependency-rich long-range code data that anchor our LCFT mixture.}

\end{enumerate}

\dscat{Assets/Code.png}{In-Repo \& Code-Adjacent Data}

\begin{enumerate}[label=\protect\numcircle{\arabic*}, leftmargin=*]

\dataset
  {OpenCoder-Annealing}
  {\hfds{OpenCoder-LLM/opc-annealing-corpus}}
  {2830910}
  {The \texttt{algorithmic\_corpus} subset of the OpenCoder~\citep{DBLP:conf/acl/HuangCLXHSXYLZC25} mid-training corpus, comprising dense, self-contained, file-level procedural code filtered from The Stack v2~\citep{DBLP:journals/corr/abs-2402-19173}.}

\dataset
  {CommonPile-GitHub}
  {\hfds{common-pile/github_archive_filtered}}
  {2288814}
  {GitHub issue reports and pull-request discussions from Common Pile~\citep{DBLP:journals/corr/abs-2506-05209}.}

\dataset
  {CraneCode}
  {\hfds{allenai/dolma3_dolmino_mix-100B-1125}}
  {1880920}
  {The \texttt{CraneCode} subset of the Olmo~3~\citep{DBLP:journals/corr/abs-2512-13961} Dolmino corpus, consisting of synthetically generated, self-contained code samples that broaden the stylistic and topical coverage of our code data.}

\dataset
  {SciCode-Domain}
  {\hfds{SciCodePile/SciCode-Domain-Code}}
  {181596}
  {Scientific-domain source code harvested from GitHub across \texttt{178} research topics (e.g., bioinformatics, numerical methods). }

\dataset
  {TheStackV2}
  {\hfds{bigcode/the-stack-v2-dedup}}
  {61111}
  {Deduplicated repository-level code contexts from the Software Heritage archive, released as part of StarCoder2~\citep{DBLP:journals/corr/abs-2402-19173}.}

\dataset
  {TheStack-Issues}
  {\hfds{bigcode/the-stack-github-issues}}
  {54985}
  {GitHub issue and pull-request conversations from The Stack~\citep{DBLP:journals/tmlr/KocetkovLALMJMF23}, providing diverse code-grounded text data.}

\end{enumerate}

\dscat{Assets/Paper.png}{Academic Papers \& PDF Data}

\begin{enumerate}[label=\protect\numcircle{\arabic*}, leftmargin=*]

\dataset
  {FinePDFs-Edu}
  {\hfds{HuggingFaceFW/finepdfs-edu}}
  {984805}
  {The education-filtered subset of FinePDFs, comprising long-form structurally rich documents extracted from openly available PDFs.}

\dataset
  {peS2o}
  {\hfds{allenai/peS2o}}
  {416910}
  {Full-text long-form research papers derived from the S2ORC~\citep{DBLP:conf/acl/LoWNKW20} corpus.}

\dataset
  {arXiv-Summarization}
  {\hfds{ccdv/arxiv-summarization}}
  {90000}
  {arXiv papers paired with their abstracts re-purposed as summaries~\citep{DBLP:conf/naacl/CohanDKBKCG18}.}

\dataset
  {Darwin-Science}
  {\hfds{GAIR/Darwin-Science}}
  {79513}
  {A quality-graded scientific document corpus, released with the Data Darwinism~\citep{DBLP:journals/corr/abs-2602-07824} study, contributing curated technical content and scientific prose.}

\end{enumerate}

\dscat{Assets/Math.png}{Math \& Math-Adjacent Data}

\begin{enumerate}[label=\protect\numcircle{\arabic*}, leftmargin=*]

\dataset
  {CraneMath}
  {\hfds{allenai/dolma3_dolmino_mix-100B-1125}}
  {1421783}
  {The \texttt{CraneMath} subset of the Olmo~3~\citep{DBLP:journals/corr/abs-2512-13961} Dolmino mixture, comprising synthetically generated mathematical text.}

\dataset
  {Nemotron-MIND}
  {\hfds{nvidia/Nemotron-MIND}}
  {657135}
  {Synthetic mathematical dialogues generated in seven conversational styles via the MIND~\citep{DBLP:conf/iclr/AkterPKSNPSC25} recipe, providing structured, math-focused multi-turn text.}

\dataset
  {NuminaMath-CoT}
  {\hfds{AI-MO/NuminaMath-CoT}}
  {433437}
  {Competition and olympiad mathematics problems paired with step-by-step solutions from the NuminaMath collection.}

\dataset
  {Nemotron-CC-Math}
  {\hfds{nvidia/Nemotron-CC-Math-v1}}
  {250470}
  {The highest quality subset of Nemotron-CC-Math~\citep{DBLP:journals/corr/abs-2508-15096}, comprised of math-heavy text from Common Crawl.}

\dataset
  {MegaMath-Web-Pro}
  {\hfds{OctoThinker/MegaMath-Web-Pro-Max}}
  {204531}
  {A high-quality, refined math web corpus released with OctoThinker~\citep{DBLP:journals/corr/abs-2506-20512}. We use it as a source of naturally occurring (rather than synthetic) mathematical prose.}

\dataset
  {Nemotron-Math-Proofs}
  {\hfds{nvidia/Nemotron-Math-Proofs-v2}}
  {99361}
  {Formal and natural-language mathematical proofs with verification and meta-verification annotations~\citep{DBLP:journals/corr/abs-2512-15489}. We strip the intermediate proof-search and verification reasoning, retaining the final proof text.}

\end{enumerate}

\dscat{Assets/General.png}{General Domain Data}

\begin{enumerate}[label=\protect\numcircle{\arabic*}, leftmargin=*]

\dataset
  {US-Patents}
  {\hfds{allenai/us-patents}}
  {361292}
  {Full-text United States patent documents sourced from PatentsView, contributing long, formally structured legal and technical writing.}

\dataset
  {Caselaw-Access}
  {\hfds{free-law/Caselaw_Access_Project}}
  {127500}
  {Digitized United States court opinions from the Caselaw Access Project, supplying long-form legal reasoning and citation-dense prose.}

\dataset
  {SEC-Filings}
  {\hfds{PleIAs/SEC}}
  {25669}
  {Form 10-K corporate filings~\citep{DBLP:journals/corr/abs-2109-14394}, providing long, highly-structured financial and regulatory documents.}

\end{enumerate}

\dscat{Assets/Book.png}{Books Data}

\begin{enumerate}[label=\protect\numcircle{\arabic*}, leftmargin=*]

\dataset
  {RedPajama-Books}
  {\hfds{togethercomputer/RedPajama-Data-1T}}
  {61206}
  {The books subset of the RedPajama~\citep{DBLP:conf/nips/WeberFAOAALNYAA24} corpus, offering full-length narrative and non-fiction prose as naturally long-form text.}

\dataset
  {BookSum}
  {\hfds{ubaada/booksum-complete-cleaned}}
  {2011}
  {Book- and chapter-level summarization pairs from BookSum~\citep{DBLP:conf/emnlp/KryscinskiRAXR22}, which couple long source documents with abstractive summaries to exercise long-range aggregation.}

\end{enumerate}

\dscat{Assets/Agentic.png}{Agentic \& Instruct-Like Data}

\begin{enumerate}[label=\protect\numcircle{\arabic*}, leftmargin=*]

\dataset
  {Magicoder-Evol-Instruct}
  {\hfds{ise-uiuc/Magicoder-Evol-Instruct-110K}}
  {910234}
  {Evolved code instruction-response pairs, decontaminated from the evol-instruct~\citep{DBLP:conf/nips/YangJWLYNP24} lineage and released with Magicoder~\citep{DBLP:conf/icml/0003W0D024}, used to inject instruction-following behavior during LCFT.}

\dataset
  {Synthetic-Text2SQL}
  {\hfds{gretelai/synthetic_text_to_sql}}
  {340000}
  {Synthetic natural-language-to-SQL pairs grounded in diverse database schemas.}

\dataset
  {SWE-Zero}
  {\hfds{AlienKevin/SWE-ZERO-12M-trajectories}}
  {299147}
  {Execution-verified software-engineering agent trajectories over real repositories.}

\dataset
  {Orca-AgentInstruct}
  {\hfds{microsoft/orca-agentinstruct-1M-v1}}
  {120000}
  {Single-turn request-response pairs obtained via the AgentInstruct~\citep{DBLP:journals/corr/abs-2407-03502} pipeline.}

\dataset
  {daVinci-Dev}
  {\hfds{GAIR/daVinci-Dev}}
  {39130}
  {The \texttt{llm\_enhanced\_prs} subset of daVinci-Dev~\citep{DBLP:journals/corr/abs-2601-18418}, comprising pull-request-derived software-development trajectories.}

\dataset
  {SynthQA}
  {\hfds{callumpurcell/SynthQA}}
  {14156}
  {Synthetic extractive question-answering examples with explanations.}

\dataset
  {TerminalTraj}
  {\hfds{m-a-p/TerminalTraj}}
  {6987}
  {Terminal-based agentic trajectories from TerminalTraj~\citep{DBLP:journals/corr/abs-2602-01244}, in which an agent issues shell commands to solve tasks.}

\end{enumerate}

\dscat{Assets/Tech.png}{Tech Content \& Forum Data}

\begin{enumerate}[label=\protect\numcircle{\arabic*}, leftmargin=*]

\dataset
  {Marin-StackExchange}
  {\hfds{marin-community/stackexchange-markdown}}
  {2000000}
  {StackExchange question-and-answer threads rendered to clean Markdown by the Marin community, covering technical forum discussion.}

\dataset
  {Coding-Tutorials}
  {\hfds{bigcode/coding_tutorials}}
  {765921}
  {Long-form programming tutorials and how-to guides, interleaving explanatory prose with code, as a source of instructional technical content.}

\end{enumerate}

\dscat{Assets/Education.png}{Educational Data}

\begin{enumerate}[label=\protect\numcircle{\arabic*}, leftmargin=*]

\dataset
  {FineWeb-Edu}
  {\hfds{HuggingFaceFW/fineweb-edu}}
  {728882}
  {The educational-heavy subset of FineWeb~\citep{DBLP:conf/nips/PenedoKALMRW024}, filtered for high-signal instructional web text.}

\dataset
  {Kaggle-Notebooks}
  {\hfds{HuggingFaceTB/issues-kaggle-notebooks}}
  {196874}
  {Structured Kaggle data-science notebooks released with StarCoder2~\citep{DBLP:journals/corr/abs-2402-19173}, interleaving code, prose explanation, and outputs into pedagogically structured long documents.}

\end{enumerate}

\subsection{Supervised Fine-Tuning (SFT) Data Mixture}
\label{subsec:sft_data_comp}

Our SFT corpus comprises $\approx$10B tokens spanning general instruction following, mathematics, code, tool use, and long-context interaction. The constituent datasets and their sample counts follow:

\begin{enumerate}[label=\protect\numcircle{\arabic*}, leftmargin=*]

\dataset
  {MegaScience}
  {\hfds{MegaScience/MegaScience}}
  {315342}
  {Large-scale scientific reasoning instruction data from MegaScience~\citep{DBLP:journals/corr/abs-2507-16812}.}

\dataset
  {Nemotron-Agentic}
  {\hfds{nvidia/Nemotron-SFT-Agentic-v2}}
  {290815}
  {The \texttt{tool\_calling} and \texttt{interactive\_agent} subsets of the Nemotron function-calling corpus, covering single-shot function calling and multi-turn interactive tool use, respectively.}

\dataset
  {LongMagpie}
  {\hfds{caskcsg/LongMagpie_multidoc_longcontext_dataset}\hsep\hfds{caskcsg/LongMagpie_singledoc_longcontext_dataset}}
  {283328}
  {The multi- and single-document long-context subsets of LongMagpie~\citep{DBLP:journals/corr/abs-2505-17134}, with synthetic long question-answering tuples.}

\dataset
  {NuminaMath-TIR}
  {\hfds{AI-MO/NuminaMath-TIR}}
  {259595}
  {Tool-integrated reasoning mathematics data in which solutions invoke a Python interpreter.}

\dataset
  {OpenCoder-SFT}
  {\hfds{OpenCoder-LLM/opc-sft-stage1}\hsep\hfds{OpenCoder-LLM/opc-sft-stage2}}
  {226597}
  {Three code instruction subsets from OpenCoder~\citep{DBLP:conf/acl/HuangCLXHSXYLZC25}: \texttt{largescale\_diverse\_instruct} and \texttt{realuser\_instruct} from stage 1, together with \texttt{package\_instruct} from stage 2, spanning diverse synthetic instructions, user-derived queries, and library-usage tasks.}

\dataset
  {Nemotron-Cascade}
  {\hfds{nvidia/Nemotron-Cascade-SFT-Stage-1}}
  {190982}
  {The \texttt{general} subset of the Nemotron-Cascade~\citep{DBLP:journals/corr/abs-2512-13607}, covering broad general-purpose instruction following.}

\dataset
  {UltraChat}
  {\hfds{openbmb/UltraChat}}
  {184827}
  {Large-scale multi-round instructional dialogues from UltraChat~\citep{DBLP:conf/emnlp/DingCXQHL0Z23}, supplying diverse open-domain conversational data.}

\dataset
  {SynSQL}
  {\hfds{seeklhy/SynSQL-2.5M}}
  {179503}
  {Synthetic text-to-SQL examples with explanations from OmniSQL~\citep{DBLP:journals/pvldb/LiWZHZJWZCSCL25}.}

\dataset
  {Dolci-Tool}
  {\hfds{allenai/Dolci-Instruct-SFT-Tool-Use}}
  {178905}
  {Tool-use instruction data from the Olmo~3~\citep{DBLP:journals/corr/abs-2512-13961} post-training suite.}

\dataset
  {MiroVerse}
  {\hfds{miromind-ai/MiroVerse-v0.1}}
  {116597}
  {Deep-research agentic trajectories from MiroThinker~\citep{DBLP:journals/corr/abs-2511-11793}.}

\dataset
  {UltraMedical}
  {\hfds{TsinghuaC3I/UltraMedical}}
  {104085}
  {Biomedical instruction data from UltraMedical~\citep{DBLP:conf/nips/ZhangZH0C0LCQZL24}, covering clinical and life-sciences question answering.}

\dataset
  {Toucan}
  {\hfds{Agent-Ark/Toucan-1.5M}}
  {89409}
  {The SFT split of Toucan~\citep{DBLP:journals/corr/abs-2510-01179}, comprising agentic trajectories on real MCPs.}

\dataset
  {WildChat}
  {\hfds{allenai/WildChat-4.8M}}
  {85766}
  {Real user--assistant conversation logs from WildChat~\citep{DBLP:conf/iclr/Zhao0HC0D24}, supplying naturally-occurring instruction distributions.}

\dataset
  {rStar-Coder}
  {\hfds{microsoft/rStar-Coder}}
  {84369}
  {Competitive-programming problems with solutions from rStar-Coder~\citep{DBLP:conf/nips/LiuZZDZSYLY25}.}

\dataset
  {Magicoder-OSS-Instruct}
  {\hfds{ise-uiuc/Magicoder-OSS-Instruct-75K}}
  {61552}
  {Open-source-derived code instruction pairs sourced via OSS-Instruct~\citep{DBLP:conf/icml/0003W0D024}.}

\dataset
  {Code-Feedback}
  {\hfds{m-a-p/Code-Feedback}}
  {60186}
  {Multi-turn code-execution-feedback dialogues from OpenCodeInterpreter~\citep{DBLP:conf/acl/ZhengZSLLFCY24}.}

\dataset
  {Magpie-Coder}
  {\hfds{Magpie-Align/Magpie-Qwen2.5-Coder-Pro-300K-v0.1}}
  {57389}
  {Synthetic coding instructions self-generated via the Magpie~\citep{DBLP:conf/iclr/XuJNDP0L25} prompting method, seeded from a code-specialized model.}

\dataset
  {ToolMind}
  {\hfds{Nanbeige/ToolMind}}
  {56963}
  {The \texttt{graph\_syn\_datasets} subset of ToolMind~\citep{DBLP:journals/corr/abs-2511-15718}, comprising function-graph-derived synthetic tool-use data.}

\dataset
  {Orca-AgentInstruct-MT}
  {\hfds{microsoft/orca-agentinstruct-1M-v1}}
  {51753}
  {The multi-turn \texttt{follow\_up} split of the AgentInstruct~\citep{DBLP:journals/corr/abs-2407-03502} data, providing extended multi-turn agentic instruction flows.}

\dataset
  {WebInstruct-CFT}
  {\hfds{TIGER-Lab/WebInstruct-CFT}}
  {49247}
  {Critique fine-tuning data from WebInstruct-CFT~\citep{DBLP:journals/corr/abs-2501-17703}, which pairs responses with model-generated reflective critiques.}

\dataset
  {NextCoder}
  {\hfds{microsoft/NextCoderDataset-Conversational}}
  {40448}
  {The conversational split of NextCoder~\citep{DBLP:conf/icml/AggarwalSA0N25}, covering multi-turn code-editing.}

\dataset
  {OpenRubrics}
  {\hfds{OpenRubrics/OpenRubric-v2}}
  {36445}
  {Rubric-conditioned instruction data from OpenRubrics~\citep{DBLP:conf/acl/LiuXYHYZW26}, pairing prompts with synthetic scoring rubrics and aligned answers.}

\dataset
  {OpenThoughts}
  {\hfds{bethgelab/CuratedThoughts}}
  {36259}
  {The \texttt{OpenThoughts-114k-Math-default} subset of the CuratedThoughts filtering of OpenThoughts~\citep{DBLP:journals/corr/abs-2506-04178}.}

\dataset
  {Jupyter-Agents}
  {\hfds{jupyter-agent/jupyter-agent-dataset}}
  {34812}
  {The \texttt{non\_thinking} subset of the Jupyter Agents data, comprising notebook-based code-execution agent trajectories.}

\dataset
  {Tulu3-Personas-IF}
  {\hfds{allenai/tulu-3-sft-personas-instruction-following}}
  {25056}
  {Persona-grounded instruction-following data from T\"ULU~3~\citep{DBLP:journals/corr/abs-2411-15124}, targeting precise constraint satisfaction.}

\dataset
  {CAMEL-Science}
  {\hfds{camel-ai/physics}\hsep\hfds{camel-ai/chemistry}\hsep\hfds{camel-ai/biology}}
  {24336}
  {Role-play-derived scientific instruction data from CAMEL~\citep{DBLP:journals/corr/abs-2303-17760}, drawn from its physics, chemistry, and biology domains.}

\dataset
  {McEval-Instruct}
  {\hfds{Multilingual-Multimodal-NLP/McEval-Instruct}}
  {22310}
  {Multilingual code instruction data released with McEval~\citep{DBLP:conf/iclr/ChaiL0YJLS0RGWW25}, spanning a broad set of programming languages.}

\dataset
  {XCoder}
  {\hfds{banksy235/XCoder-80K}}
  {20811}
  {A quality- and diversity-selected code instruction set from XCoder~\citep{DBLP:journals/corr/abs-2409-03810}.}

\dataset
  {Research-Plan-Gen}
  {\hfds{facebook/research-plan-gen}}
  {19389}
  {Research-planning instruction data released alongside work on rubric-rewarded AI co-scientists~\citep{DBLP:journals/corr/abs-2512-23707}, pairing research goals with structured plans.}

\dataset
  {Mini-SWE}
  {\hfds{ricdomolm/mini-coder-trajs-400k}}
  {17715}
  {Execution-verified \texttt{mini-swe-agent}~\citep{DBLP:conf/nips/YangJWLYNP24} rollouts.}

\dataset
  {RAGBench}
  {\hfds{galileo-ai/ragbench}}
  {17084}
  {Examples from RAGBench~\citep{DBLP:journals/corr/abs-2407-11005}, pairing retrieved context with grounded answers to reinforce long-context grounding.}

\dataset
  {Nemotron-IF}
  {\hfds{nvidia/Nemotron-Instruction-Following-Chat-v1}}
  {12002}
  {Chat-formatted instruction-following data emphasizing precise constraint adherence.}

\dataset
  {Nemotron-Terminal}
  {\hfds{nvidia/Nemotron-Terminal-Corpus}}
  {10740}
  {The \texttt{skill\_based\_medium} subset of the Nemotron Terminal corpus~\citep{DBLP:journals/corr/abs-2602-21193}, comprising terminal-agent trajectories.}

\dataset
  {LongMIT}
  {\hfds{donmaclean/LongMIT-128K}}
  {9425}
  {Long-context multi-hop instruction data (up to 128K tokens) constructed via the best-practice recipe of~\citet{DBLP:conf/acl/ChenC0GLZ00L25}.}

\dataset
  {CodeIO-PyEdu}
  {\hfds{hkust-nlp/CodeIO-PyEdu-Reasoning}}
  {9324}
  {Data from CodeIO~\citep{DBLP:conf/icml/LiGYX0H25}, which gleans reasoning patterns from single- and multi-turn code input-output prediction tasks.}

\dataset
  {LongAlign}
  {\hfds{zai-org/LongAlign-10k}}
  {5878}
  {Long-context alignment instructions from LongAlign~\citep{DBLP:conf/emnlp/BaiLZHQH0DL24}, designed to align models on tasks with long inputs.}

\dataset
  {LiteCoder-Terminal}
  {\hfds{Lite-Coder/LiteCoder-Terminal-SFT}}
  {5407}
  {Long-horizon terminal agent trajectories from LiteCoder-Terminal~\citep{DBLP:journals/corr/abs-2605-29559}.}

\dataset
  {When2Call}
  {\hfds{nvidia/When2Call}}
  {4528}
  {Irrelevant tool-calling abstention training data from When2Call~\citep{DBLP:conf/naacl/RossMS25}.}

\dataset
  {ToolMind-Web}
  {\hfds{Nanbeige/ToolMind-Web-QA}}
  {3392}
  {Web-search tool-use question-answering data released with Nanbeige4.1~\citep{DBLP:journals/corr/abs-2602-13367}.}

\dataset
  {LongWriter}
  {\hfds{zai-org/LongWriter-6k}}
  {2304}
  {Ultra-long-form generation instructions from LongWriter~\citep{DBLP:conf/iclr/BaiZLZZ0D0L25}, targeting coherent outputs exceeding 10K words.}

\dataset
  {ToolACE}
  {\hfds{Team-ACE/ToolACE}}
  {2094}
  {Function-calling data from ToolACE~\citep{DBLP:conf/iclr/Liu0ZHYL0GLY0WN25}, emphasizing diverse API invocation.}

\dataset
  {APIGen-MT}
  {\hfds{Salesforce/APIGen-MT-5k}}
  {410}
  {Multi-turn agent--human interaction data synthesized via APIGen-MT~\citep{DBLP:conf/nips/PrabhakarLZZAWL25}.}

\dataset
  {HardGen}
  {\hfds{Bingguang/HardGen}}
  {188}
  {Hard tool-use samples generated for failure-driven learning of tool agents~\citep{DBLP:journals/corr/abs-2601-01498}.}

\dataset
  {Goedel-Prover}
  {\hfds{Goedel-LM/SFT_dataset_v2}}
  {179}
  {Formal theorem-proving data from Goedel-Prover-V2~\citep{DBLP:journals/corr/abs-2508-03613}.}

\end{enumerate}

\onecolumn
\raggedbottom
\section{Detailed Results}
\label{sec:appendix_detailed_res}

\begingroup
\setlength{\intextsep}{3pt plus 1pt minus 1pt}
\setlength{\textfloatsep}{3pt plus 1pt minus 1pt}
\setlength{\abovecaptionskip}{2.1pt}
\setlength{\belowcaptionskip}{0pt}
\begin{table}[H]
\centering
\scalebox{0.8}{\footnotesize \gcode\;\textbf{\texttt{Code-Specialised}} \qquad \gcext\;\textbf{\texttt{Context-Extended-After-Post-Training}} \qquad \gpost\;\textbf{\texttt{Preference/RL-Tuned}}}
\scalebox{0.55}{%
\setlength{\tabcolsep}{4pt}%
\begin{tabular}{@{}c l l r r rr rr rr@{}}
\toprule
& \multirow{3}{*}{\textbf{\texttt{Model}}} & & \multirow{3}{*}{\textbf{\texttt{Size}}} & \multicolumn{1}{c}{\multirow{3}{*}{\shortstack{\textbf{\texttt{Max}}\\ \textbf{\texttt{Length}}}}} & \multicolumn{2}{c}{\textbf{\texttt{LooGLE V2 Code}}} & \multicolumn{2}{c}{\textbf{\texttt{LongCodeQA}}} & \multicolumn{2}{c}{\textbf{\texttt{RepoQA}}} \\
\cmidrule(lr){6-7} \cmidrule(lr){8-9} \cmidrule(lr){10-11}
& & & & & \hc{\textbf{\texttt{Call}}} & \hc{\textbf{\texttt{Version}}} & \hc{\textbf{\texttt{64K}}} & \hc{\textbf{\texttt{128K}}} & \hc{\textbf{\texttt{64K}}} & \hc{\textbf{\texttt{128K}}} \\
& & & & & \hc{\textbf{\texttt{Graph}}} & \hc{\textbf{\texttt{Control}}} & & & & \\
\midrule
\multirow{1}{*}{\rotatebox[origin=c]{90}{\textbf{\texttt{XS}}}}
& \textbf{\hsq{\texttt{OctoLong-0.6B-Instruct}}} & \gcode & \texttt{0.6B} & \texttt{128K} & \olsc{6.04} & \olsc{7.69} & \olsc{17.88} & \olsc{20.74} & \olsc{7.12} & \olsc{5.11} \\
\midrule
\multirow{4}{*}{\rotatebox[origin=c]{90}{\textbf{\texttt{S}}}}
& \rmodel{meta-llama/Llama-3.2-1B-Instruct} & \gpost & \texttt{1B} & \texttt{128K} & \olsc{0.84} & \olsc{8.82} & \olsc{22.34} & \olsc{9.47} & \olsc{0.86} & \olsc{0.44} \\
& \rmodel{01-ai/Yi-Coder-1.5B-Chat} & \gcode\;\gpost & \texttt{1.5B} & \texttt{128K} & \olsc{2.51} & \olsc{4.13} & \olsc{2.86} & \olsc{1.96} & \olsc{0.93} & \olsc{0.70} \\
& \textbf{\hsq{\texttt{OctoLong-1.7B-Instruct}}} & \gcode & \texttt{1.7B} & \texttt{128K} & \olsc{14.51} & \olsc{15.68} & \olsc{48.69} & \olsc{40.93} & \olsc{27.41} & \olsc{24.22} \\
& \rmodel{ibm-granite/granite-3.1-2b-instruct} & \gpost & \texttt{2B} & \texttt{128K} & \olsc{18.71} & \olsc{11.53} & \olsc{38.51} & \olsc{33.79} & \olsc{31.24} & \olsc{18.59} \\
\midrule
\multirow{4}{*}{\rotatebox[origin=c]{90}{\textbf{\texttt{M}}}}
& \rmodel{meta-llama/Llama-3.2-3B-Instruct} & \gpost & \texttt{3B} & \texttt{128K} & \olsc{28.32} & \olsc{19.48} & \olsc{23.42} & \olsc{14.15} & \olsc{7.57} & \olsc{5.24} \\
& \rmodel{Qwen/Qwen3-4B-Instruct-2507} & \gpost & \texttt{4B} & \texttt{256K} & \olsc{27.81} & \olsc{29.11} & \olsc{63.38} & \olsc{63.90} & \olsc{60.42} & \olsc{54.37} \\
& \textbf{\hsq{\texttt{OctoLong-4B-Instruct}}} & \gcode & \texttt{4B} & \texttt{128K} & \olsc{38.58} & \olsc{21.04} & \olsc{68.82} & \olsc{59.37} & \olsc{63.18} & \olsc{52.96} \\
& \rmodel{arcee-ai/AFM-4.5B} & \gpost & \texttt{4.5B} & \texttt{64K} & \olsc{24.31} & \olsc{7.95} & \olsc{61.02} & \texttt{-} & \olsc{6.96} & \texttt{-} \\
\midrule
\multirow{12}{*}{\rotatebox[origin=c]{90}{\textbf{\texttt{L}}}}
& \rmodel{Qwen/Qwen2.5-7B-Instruct-1M} & \gpost & \texttt{7B} & \texttt{1024K} & \olsc{34.20} & \olsc{19.62} & \olsc{58.41} & \olsc{63.51} & \olsc{45.27} & \olsc{41.64} \\
& \rmodel{aws-prototyping/MegaBeam-Mistral-7B-512k} & \gcext & \texttt{7B} & \texttt{512K} & \olsc{26.13} & \olsc{10.19} & \olsc{38.23} & \olsc{42.21} & \olsc{32.11} & \olsc{20.33} \\
& \rmodel{internlm/internlm2_5-7b-chat-1m} & \gpost & \texttt{7B} & \texttt{1024K} & \olsc{8.79} & \olsc{18.01} & \olsc{39.29} & \olsc{40.23} & \olsc{40.02} & \olsc{35.46} \\
& \textbf{\hsq{\texttt{OctoLong-8B-Instruct}}} & \gcode & \texttt{8B} & \texttt{128K} & \olsc{33.19} & \olsc{26.59} & \olsc{65.52} & \olsc{64.85} & \olsc{67.87} & \olsc{61.67} \\
& \rmodel{meta-llama/Llama-3.1-8B-Instruct} & \gpost & \texttt{8B} & \texttt{128K} & \olsc{22.79} & \olsc{24.01} & \olsc{55.72} & \olsc{47.92} & \olsc{37.51} & \olsc{32.18} \\
& \rmodel{nvidia/Llama-3.1-Nemotron-8B-UltraLong-1M-Instruct} & \gcext & \texttt{8B} & \texttt{1024K} & \olsc{23.11} & \olsc{19.61} & \olsc{36.55} & \olsc{34.29} & \olsc{31.62} & \olsc{22.48} \\
& \rmodel{princeton-nlp/Llama-3-8B-ProLong-512k-Instruct} & \gcext & \texttt{8B} & \texttt{512K} & \olsc{21.26} & \olsc{16.29} & \olsc{27.14} & \olsc{24.58} & \olsc{38.51} & \olsc{32.67} \\
& \rmodel{ibm-granite/granite-3.1-8b-instruct} & \gpost & \texttt{8B} & \texttt{128K} & \olsc{27.52} & \olsc{16.94} & \olsc{33.52} & \olsc{29.61} & \olsc{44.16} & \olsc{31.83} \\
& \rmodel{gradientai/Llama-3-8B-Instruct-262k} & \gcext & \texttt{8B} & \texttt{256K} & \olsc{6.39} & \olsc{15.53} & \olsc{9.34} & \olsc{4.89} & \olsc{31.61} & \olsc{30.09} \\
& \rmodel{mistralai/Ministral-8B-Instruct-2410} & \gpost & \texttt{8B} & \texttt{128K} & \olsc{17.26} & \olsc{19.51} & \olsc{55.89} & \olsc{48.05} & \olsc{27.55} & \olsc{16.11} \\
& \rmodel{zai-org/glm-4-9b-chat-1m} & \gpost & \texttt{9B} & \texttt{1024K} & \olsc{23.55} & \olsc{21.64} & \olsc{66.01} & \olsc{57.95} & \olsc{61.58} & \olsc{56.70} \\
& \rmodel{01-ai/Yi-Coder-9B-Chat} & \gcode\;\gpost & \texttt{9B} & \texttt{128K} & \olsc{26.33} & \olsc{7.89} & \olsc{9.73} & \olsc{3.63} & \olsc{16.74} & \olsc{8.89} \\
\midrule
\multirow{2}{*}{\rotatebox[origin=c]{90}{\textbf{\texttt{XL}}}}
& \textbf{\hsq{\texttt{OctoLong-14B-Instruct}}} & \gcode & \texttt{14B} & \texttt{128K} & \olsc{45.87} & \olsc{33.86} & \olsc{78.62} & \olsc{77.41} & \olsc{80.58} & \olsc{71.44} \\
& \rmodel{Qwen/Qwen2.5-14B-Instruct-1M} & \gpost & \texttt{14B} & \texttt{1024K} & \olsc{43.25} & \olsc{31.10} & \olsc{73.23} & \olsc{70.97} & \olsc{76.83} & \olsc{73.28} \\
\midrule
\multirow{4}{*}{\rotatebox[origin=c]{90}{\textbf{\texttt{Abl.}}}}
& \multirow{2}{*}{\textbf{\hsq{\texttt{OctoLong-8B-Instruct - Cross-Repo Code}}}} & \multirow{2}{*}{\gcode} & \multirow{2}{*}{\texttt{8B}} & \multirow{2}{*}{\texttt{128K}} & \hc{\texttt{28.78}} & \hc{\texttt{23.98}} & \hc{\texttt{64.11}} & \hc{\texttt{60.23}} & \hc{\texttt{64.24}} & \hc{\texttt{58.91}} \\
\addlinespace[-0.2em]
& & & & & \hc{\diffdo{4.41}} & \hc{\diffdo{2.61}} & \hc{\diffdo{1.41}} & \hc{\diffdo{4.62}} & \hc{\diffdo{3.63}} & \hc{\diffdo{2.76}} \\
\addlinespace[0.12em]
& \multirow{2}{*}{\textbf{\hsq{\texttt{OctoLong-8B-Instruct - Model Merging}}}} & \multirow{2}{*}{\gcode} & \multirow{2}{*}{\texttt{8B}} & \multirow{2}{*}{\texttt{128K}} & \hc{\texttt{30.55}} & \hc{\texttt{25.70}} & \hc{\texttt{64.60}} & \hc{\texttt{63.89}} & \hc{\texttt{66.39}} & \hc{\texttt{61.07}} \\
\addlinespace[-0.2em]
& & & & & \hc{\diffdo{2.64}} & \hc{\diffdo{0.89}} & \hc{\diffdo{0.92}} & \hc{\diffdo{0.96}} & \hc{\diffdo{1.48}} & \hc{\diffdo{0.60}} \\
\addlinespace[0.12em]
\bottomrule
\end{tabular}%
}
\caption{Detailed per-setting scores on the coding long-context evaluation suite (\textbf{\texttt{RQ1}}). \texttt{LooGLE V2 Code}~\citep{DBLP:journals/corr/abs-2510-22548} is split into its call-graph and version-control subsets, while \texttt{LongCodeQA}~\citep{DBLP:journals/corr/abs-2505-07897} and \texttt{RepoQA}~\citep{DBLP:journals/corr/abs-2406-06025} are reported at their \texttt{64K} and \texttt{128K} context settings. We note the strong performance of the \texttt{OctoLong-Instruct} suite across the retrieval- and state-tracking-heavy multi-file and repository-level coding tasks. \cref{subsec:RQ1} discusses the results in detail.}
\label{tab:Results_RQ1}
\end{table}
\begin{table}[H]
\centering
\scalebox{0.8}{\footnotesize \gcode\;\textbf{\texttt{Code-Specialised}} \qquad \gcext\;\textbf{\texttt{Context-Extended-After-Post-Training}} \qquad \gpost\;\textbf{\texttt{Preference/RL-Tuned}}}
\scalebox{0.55}{%
\setlength{\tabcolsep}{4pt}%
\begin{tabular}{@{}c l l r r rrrr rr r@{}}
\toprule
& \multirow{3}{*}{\textbf{\texttt{Model}}} & & \multirow{3}{*}{\textbf{\texttt{Size}}} & \multicolumn{1}{c}{\multirow{3}{*}{\shortstack{\textbf{\texttt{Max}}\\ \textbf{\texttt{Length}}}}} & \multicolumn{4}{c}{\textbf{\texttt{LongBench V2 (Medium)}}} & \multicolumn{2}{c}{\textbf{\texttt{OpenAI MRCR}}} & \multicolumn{1}{c}{\multirow{3}{*}{\shortstack{\textbf{\texttt{AA-}}\\ \textbf{\texttt{LCR}}}}} \\
\cmidrule(lr){6-9} \cmidrule(lr){10-11}
& & & & & \hc{\textbf{\texttt{Single}}} & \hc{\textbf{\texttt{Multi}}} & \hc{\textbf{\texttt{Long}}} & \hc{\textbf{\texttt{Long}}} & \hc{\textbf{\texttt{32K--}}} & \hc{\textbf{\texttt{64K--}}} & \\
& & & & & \hc{\textbf{\texttt{Doc}}} & \hc{\textbf{\texttt{Doc}}} & \hc{\textbf{\texttt{ICL}}} & \hc{\textbf{\texttt{Dialogue}}} & \hc{\textbf{\texttt{64K}}} & \hc{\textbf{\texttt{128K}}} & \\
\midrule
\multirow{1}{*}{\rotatebox[origin=c]{90}{\textbf{\texttt{XS}}}}
& \textbf{\hsq{\texttt{OctoLong-0.6B-Instruct}}} & \gcode & \texttt{0.6B} & \texttt{128K} & \olsc{14.81} & \olsc{26.52} & \olsc{22.18} & \olsc{27.76} & \olsc{12.89} & \olsc{5.54} & \ollcr{5.00} \\
\midrule
\multirow{4}{*}{\rotatebox[origin=c]{90}{\textbf{\texttt{S}}}}
& \rmodel{meta-llama/Llama-3.2-1B-Instruct} & \gpost & \texttt{1B} & \texttt{128K} & \olsc{26.28} & \olsc{12.94} & \olsc{20.35} & \olsc{24.88} & \olsc{3.14} & \olsc{1.92} & \ollcr{7.00} \\
& \rmodel{01-ai/Yi-Coder-1.5B-Chat} & \gcode\;\gpost & \texttt{1.5B} & \texttt{128K} & \olsc{18.53} & \olsc{15.64} & \olsc{11.27} & \olsc{20.50} & \olsc{1.18} & \olsc{0.00} & \ollcr{1.00} \\
& \textbf{\hsq{\texttt{OctoLong-1.7B-Instruct}}} & \gcode & \texttt{1.7B} & \texttt{128K} & \olsc{25.09} & \olsc{27.24} & \olsc{17.24} & \olsc{36.39} & \olsc{22.79} & \olsc{16.27} & \ollcr{8.00} \\
& \rmodel{ibm-granite/granite-3.1-2b-instruct} & \gpost & \texttt{2B} & \texttt{128K} & \olsc{24.21} & \olsc{22.16} & \olsc{25.12} & \olsc{26.20} & \olsc{18.25} & \olsc{8.94} & \ollcr{6.00} \\
\midrule
\multirow{4}{*}{\rotatebox[origin=c]{90}{\textbf{\texttt{M}}}}
& \rmodel{meta-llama/Llama-3.2-3B-Instruct} & \gpost & \texttt{3B} & \texttt{128K} & \olsc{19.47} & \olsc{20.84} & \olsc{25.04} & \olsc{24.75} & \olsc{22.07} & \olsc{8.97} & \ollcr{7.00} \\
& \rmodel{Qwen/Qwen3-4B-Instruct-2507} & \gpost & \texttt{4B} & \texttt{256K} & \olsc{26.32} & \olsc{27.51} & \olsc{21.72} & \olsc{43.24} & \olsc{39.74} & \olsc{37.42} & \ollcr{10.00} \\
& \textbf{\hsq{\texttt{OctoLong-4B-Instruct}}} & \gcode & \texttt{4B} & \texttt{128K} & \olsc{23.36} & \olsc{24.90} & \olsc{23.19} & \olsc{41.34} & \olsc{38.45} & \olsc{35.37} & \ollcr{10.00} \\
& \rmodel{arcee-ai/AFM-4.5B} & \gpost & \texttt{4.5B} & \texttt{64K} & \olsc{17.44} & \olsc{25.56} & \olsc{20.85} & \olsc{26.22} & \olsc{5.91} & \texttt{-} & \ollcr{7.00} \\
\midrule
\multirow{12}{*}{\rotatebox[origin=c]{90}{\textbf{\texttt{L}}}}
& \rmodel{Qwen/Qwen2.5-7B-Instruct-1M} & \gpost & \texttt{7B} & \texttt{1024K} & \olsc{33.21} & \olsc{29.19} & \olsc{28.42} & \olsc{22.14} & \olsc{36.41} & \olsc{34.19} & \ollcr{13.00} \\
& \rmodel{aws-prototyping/MegaBeam-Mistral-7B-512k} & \gcext & \texttt{7B} & \texttt{512K} & \olsc{20.88} & \olsc{25.04} & \olsc{21.85} & \olsc{26.48} & \olsc{0.00} & \olsc{0.00} & \ollcr{6.00} \\
& \rmodel{internlm/internlm2_5-7b-chat-1m} & \gpost & \texttt{7B} & \texttt{1024K} & \olsc{19.25} & \olsc{25.44} & \olsc{15.11} & \olsc{41.41} & \olsc{30.49} & \olsc{24.55} & \ollcr{6.00} \\
& \textbf{\hsq{\texttt{OctoLong-8B-Instruct}}} & \gcode & \texttt{8B} & \texttt{128K} & \olsc{28.47} & \olsc{36.14} & \olsc{34.89} & \olsc{57.22} & \olsc{42.78} & \olsc{38.96} & \ollcr{13.00} \\
& \rmodel{meta-llama/Llama-3.1-8B-Instruct} & \gpost & \texttt{8B} & \texttt{128K} & \olsc{32.59} & \olsc{25.71} & \olsc{32.66} & \olsc{21.69} & \olsc{35.67} & \olsc{25.58} & \ollcr{13.00} \\
& \rmodel{nvidia/Llama-3.1-Nemotron-8B-UltraLong-1M-Instruct} & \gcext & \texttt{8B} & \texttt{1024K} & \olsc{28.44} & \olsc{20.57} & \olsc{26.45} & \olsc{21.33} & \olsc{4.90} & \olsc{3.44} & \ollcr{10.00} \\
& \rmodel{princeton-nlp/Llama-3-8B-ProLong-512k-Instruct} & \gcext & \texttt{8B} & \texttt{512K} & \olsc{29.25} & \olsc{26.86} & \olsc{26.79} & \olsc{15.97} & \olsc{43.33} & \olsc{40.08} & \ollcr{6.00} \\
& \rmodel{ibm-granite/granite-3.1-8b-instruct} & \gpost & \texttt{8B} & \texttt{128K} & \olsc{28.38} & \olsc{29.51} & \olsc{16.17} & \olsc{15.89} & \olsc{29.18} & \olsc{27.44} & \ollcr{7.00} \\
& \rmodel{gradientai/Llama-3-8B-Instruct-262k} & \gcext & \texttt{8B} & \texttt{256K} & \olsc{27.54} & \olsc{32.67} & \olsc{20.23} & \olsc{53.44} & \olsc{41.78} & \olsc{23.88} & \ollcr{2.00} \\
& \rmodel{mistralai/Ministral-8B-Instruct-2410} & \gpost & \texttt{8B} & \texttt{128K} & \olsc{31.33} & \olsc{26.48} & \olsc{22.74} & \olsc{21.67} & \olsc{21.59} & \olsc{4.72} & \ollcr{4.00} \\
& \rmodel{zai-org/glm-4-9b-chat-1m} & \gpost & \texttt{9B} & \texttt{1024K} & \olsc{26.20} & \olsc{25.33} & \olsc{11.71} & \olsc{51.56} & \olsc{4.29} & \olsc{0.00} & \ollcr{12.00} \\
& \rmodel{01-ai/Yi-Coder-9B-Chat} & \gcode\;\gpost & \texttt{9B} & \texttt{128K} & \olsc{19.44} & \olsc{21.52} & \olsc{18.26} & \olsc{26.22} & \olsc{2.55} & \olsc{1.22} & \ollcr{5.00} \\
\midrule
\multirow{2}{*}{\rotatebox[origin=c]{90}{\textbf{\texttt{XL}}}}
& \textbf{\hsq{\texttt{OctoLong-14B-Instruct}}} & \gcode & \texttt{14B} & \texttt{128K} & \olsc{37.28} & \olsc{35.44} & \olsc{30.48} & \olsc{62.87} & \olsc{46.93} & \olsc{42.81} & \ollcr{21.00} \\
& \rmodel{Qwen/Qwen2.5-14B-Instruct-1M} & \gpost & \texttt{14B} & \texttt{1024K} & \olsc{29.91} & \olsc{30.74} & \olsc{33.23} & \olsc{31.88} & \olsc{37.49} & \olsc{35.76} & \ollcr{23.00} \\
\midrule
\multirow{4}{*}{\rotatebox[origin=c]{90}{\textbf{\texttt{Abl.}}}}
& \multirow{2}{*}{\textbf{\hsq{\texttt{OctoLong-8B-Instruct - Cross-Repo Code}}}} & \multirow{2}{*}{\gcode} & \multirow{2}{*}{\texttt{8B}} & \multirow{2}{*}{\texttt{128K}} & \hc{\texttt{27.76}} & \hc{\texttt{35.68}} & \hc{\texttt{31.77}} & \hc{\texttt{52.89}} & \hc{\texttt{39.44}} & \hc{\texttt{34.31}} & \hc{\texttt{11.00}} \\
\addlinespace[-0.2em]
& & & & & \hc{\diffdo{0.71}} & \hc{\diffdo{0.46}} & \hc{\diffdo{3.12}} & \hc{\diffdo{4.33}} & \hc{\diffdo{3.34}} & \hc{\diffdo{4.65}} & \hc{\diffdo{2.00}} \\
\addlinespace[0.12em]
& \multirow{2}{*}{\textbf{\hsq{\texttt{OctoLong-8B-Instruct - Model Merging}}}} & \multirow{2}{*}{\gcode} & \multirow{2}{*}{\texttt{8B}} & \multirow{2}{*}{\texttt{128K}} & \hc{\texttt{28.52}} & \hc{\texttt{35.97}} & \hc{\texttt{34.06}} & \hc{\texttt{55.88}} & \hc{\texttt{41.58}} & \hc{\texttt{36.82}} & \hc{\texttt{12.00}} \\
\addlinespace[-0.2em]
& & & & & \hc{\diffup{0.05}} & \hc{\diffdo{0.17}} & \hc{\diffdo{0.83}} & \hc{\diffdo{1.34}} & \hc{\diffdo{1.20}} & \hc{\diffdo{2.14}} & \hc{\diffdo{1.00}} \\
\addlinespace[0.12em]
\bottomrule
\end{tabular}%
}
\caption{Detailed per-setting scores on the general-domain long-context evaluation suite (\textbf{\texttt{RQ2}}). \texttt{LongBench V2 Medium}~\citep{DBLP:conf/acl/BaiTZ0WLCX0D0L25} is split into its single-document, multi-document, long in-context learning, and long-dialogue subsets; \texttt{OpenAI MRCR}~\citep{openai2025mrcr} is reported at its \texttt{32K--64K} and \texttt{64K--128K} depth bins; \texttt{AA-LCR}~\citep{artificialanalysis2025lcr} denotes the Artificial Analysis long-context reasoning score. The \texttt{OctoLong-Instruct} suite's consistently strong performance across a range of realistic and industry-standard long-context benchmarks is noteworthy. A detailed discussion of results resides in \cref{subsec:RQ2}.}
\label{tab:Results_RQ2}
\end{table}
\begin{table}[H]
\centering
\scalebox{0.8}{\footnotesize \gcode\;\textbf{\texttt{Code-Specialised}} \qquad \gcext\;\textbf{\texttt{Context-Extended-After-Post-Training}} \qquad \gpost\;\textbf{\texttt{Preference/RL-Tuned}}}
\scalebox{0.55}{%
\setlength{\tabcolsep}{4pt}%
\begin{tabular}{@{}c l l r r rrr rrr@{}}
\toprule
& \multirow{3}{*}{\textbf{\texttt{Model}}} & & \multirow{3}{*}{\textbf{\texttt{Size}}} & \multicolumn{1}{c}{\multirow{3}{*}{\shortstack{\textbf{\texttt{Max}}\\ \textbf{\texttt{Length}}}}} & \multicolumn{3}{c}{\textbf{\texttt{BFCL V3}}} & \multicolumn{3}{c}{\textbf{\texttt{$\tau^2$-Bench}}} \\
\cmidrule(lr){6-8} \cmidrule(lr){9-11}
& & & & & \hc{\textbf{\texttt{Non-}}} & \hc{\textbf{\texttt{Live}}} & \hc{\textbf{\texttt{Multi-}}} & \hc{\textbf{\texttt{Air-}}} & \hc{\textbf{\texttt{Re-}}} & \hc{\textbf{\texttt{Tele-}}} \\
& & & & & \hc{\textbf{\texttt{Live}}} & & \hc{\textbf{\texttt{Turn}}} & \hc{\textbf{\texttt{line}}} & \hc{\textbf{\texttt{tail}}} & \hc{\textbf{\texttt{com}}} \\
\midrule
\multirow{1}{*}{\rotatebox[origin=c]{90}{\textbf{\texttt{XS}}}}
& \textbf{\hsq{\texttt{OctoLong-0.6B-Instruct}}} & \gcode & \texttt{0.6B} & \texttt{128K} & \olsc{63.66} & \olsc{53.42} & \olsc{4.31} & \olsc{32.14} & \olsc{7.63} & \olsc{7.89} \\
\midrule
\multirow{4}{*}{\rotatebox[origin=c]{90}{\textbf{\texttt{S}}}}
& \rmodel{meta-llama/Llama-3.2-1B-Instruct} & \gpost & \texttt{1B} & \texttt{128K} & \olsc{36.74} & \olsc{22.43} & \olsc{0.00} & \olsc{30.22} & \olsc{2.81} & \olsc{7.88} \\
& \rmodel{01-ai/Yi-Coder-1.5B-Chat} & \gcode\;\gpost & \texttt{1.5B} & \texttt{128K} & \olsc{25.21} & \olsc{19.47} & \olsc{0.00} & \olsc{43.52} & \olsc{4.67} & \olsc{3.82} \\
& \textbf{\hsq{\texttt{OctoLong-1.7B-Instruct}}} & \gcode & \texttt{1.7B} & \texttt{128K} & \olsc{74.72} & \olsc{63.19} & \olsc{6.84} & \olsc{26.91} & \olsc{11.42} & \olsc{12.71} \\
& \rmodel{ibm-granite/granite-3.1-2b-instruct} & \gpost & \texttt{2B} & \texttt{128K} & \olsc{67.79} & \olsc{55.38} & \olsc{2.41} & \olsc{38.68} & \olsc{6.41} & \olsc{8.72} \\
\midrule
\multirow{4}{*}{\rotatebox[origin=c]{90}{\textbf{\texttt{M}}}}
& \rmodel{meta-llama/Llama-3.2-3B-Instruct} & \gpost & \texttt{3B} & \texttt{128K} & \olsc{38.52} & \olsc{29.64} & \olsc{4.49} & \olsc{28.46} & \olsc{9.02} & \olsc{6.41} \\
& \rmodel{Qwen/Qwen3-4B-Instruct-2507} & \gpost & \texttt{4B} & \texttt{256K} & \olsc{81.37} & \olsc{70.48} & \olsc{21.57} & \olsc{34.92} & \olsc{41.97} & \olsc{15.67} \\
& \textbf{\hsq{\texttt{OctoLong-4B-Instruct}}} & \gcode & \texttt{4B} & \texttt{128K} & \olsc{81.49} & \olsc{69.14} & \olsc{24.72} & \olsc{33.81} & \olsc{26.11} & \olsc{15.72} \\
& \rmodel{arcee-ai/AFM-4.5B} & \gpost & \texttt{4.5B} & \texttt{64K} & \olsc{72.34} & \olsc{54.07} & \olsc{3.45} & \olsc{43.31} & \olsc{7.87} & \olsc{9.26} \\
\midrule
\multirow{12}{*}{\rotatebox[origin=c]{90}{\textbf{\texttt{L}}}}
& \rmodel{Qwen/Qwen2.5-7B-Instruct-1M} & \gpost & \texttt{7B} & \texttt{1024K} & \olsc{71.17} & \olsc{66.42} & \olsc{16.11} & \olsc{29.41} & \olsc{22.76} & \olsc{16.88} \\
& \rmodel{aws-prototyping/MegaBeam-Mistral-7B-512k} & \gcext & \texttt{7B} & \texttt{512K} & \olsc{35.27} & \olsc{27.61} & \olsc{0.81} & \olsc{0.00} & \olsc{0.00} & \olsc{0.00} \\
& \rmodel{internlm/internlm2_5-7b-chat-1m} & \gpost & \texttt{7B} & \texttt{1024K} & \olsc{42.61} & \olsc{40.61} & \olsc{1.46} & \olsc{48.31} & \olsc{7.24} & \olsc{11.59} \\
& \textbf{\hsq{\texttt{OctoLong-8B-Instruct}}} & \gcode & \texttt{8B} & \texttt{128K} & \olsc{82.47} & \olsc{72.94} & \olsc{29.59} & \olsc{30.62} & \olsc{34.81} & \olsc{16.61} \\
& \rmodel{meta-llama/Llama-3.1-8B-Instruct} & \gpost & \texttt{8B} & \texttt{128K} & \olsc{42.79} & \olsc{39.98} & \olsc{10.21} & \olsc{33.16} & \olsc{9.19} & \olsc{8.41} \\
& \rmodel{nvidia/Llama-3.1-Nemotron-8B-UltraLong-1M-Instruct} & \gcext & \texttt{8B} & \texttt{1024K} & \olsc{41.26} & \olsc{38.11} & \olsc{1.57} & \olsc{40.47} & \olsc{5.23} & \olsc{4.91} \\
& \rmodel{princeton-nlp/Llama-3-8B-ProLong-512k-Instruct} & \gcext & \texttt{8B} & \texttt{512K} & \olsc{60.11} & \olsc{53.41} & \olsc{3.48} & \olsc{41.16} & \olsc{10.28} & \olsc{11.77} \\
& \rmodel{ibm-granite/granite-3.1-8b-instruct} & \gpost & \texttt{8B} & \texttt{128K} & \olsc{79.69} & \olsc{59.81} & \olsc{7.82} & \olsc{46.51} & \olsc{6.44} & \olsc{10.31} \\
& \rmodel{gradientai/Llama-3-8B-Instruct-262k} & \gcext & \texttt{8B} & \texttt{256K} & \olsc{25.23} & \olsc{31.55} & \olsc{3.46} & \olsc{41.43} & \olsc{7.06} & \olsc{9.22} \\
& \rmodel{mistralai/Ministral-8B-Instruct-2410} & \gpost & \texttt{8B} & \texttt{128K} & \olsc{14.43} & \olsc{16.71} & \olsc{0.00} & \olsc{20.43} & \olsc{2.41} & \olsc{4.08} \\
& \rmodel{zai-org/glm-4-9b-chat-1m} & \gpost & \texttt{9B} & \texttt{1024K} & \olsc{54.76} & \olsc{44.04} & \olsc{2.81} & \olsc{46.60} & \olsc{6.91} & \olsc{11.38} \\
& \rmodel{01-ai/Yi-Coder-9B-Chat} & \gcode\;\gpost & \texttt{9B} & \texttt{128K} & \olsc{70.57} & \olsc{68.79} & \olsc{10.36} & \olsc{45.66} & \olsc{6.24} & \olsc{10.43} \\
\midrule
\multirow{2}{*}{\rotatebox[origin=c]{90}{\textbf{\texttt{XL}}}}
& \textbf{\hsq{\texttt{OctoLong-14B-Instruct}}} & \gcode & \texttt{14B} & \texttt{128K} & \olsc{84.41} & \olsc{75.91} & \olsc{41.82} & \olsc{45.48} & \olsc{43.95} & \olsc{14.68} \\
& \rmodel{Qwen/Qwen2.5-14B-Instruct-1M} & \gpost & \texttt{14B} & \texttt{1024K} & \olsc{72.24} & \olsc{69.17} & \olsc{26.01} & \olsc{27.61} & \olsc{30.14} & \olsc{20.62} \\
\midrule
\multirow{4}{*}{\rotatebox[origin=c]{90}{\textbf{\texttt{Abl.}}}}
& \multirow{2}{*}{\textbf{\hsq{\texttt{OctoLong-8B-Instruct - Cross-Repo Code}}}} & \multirow{2}{*}{\gcode} & \multirow{2}{*}{\texttt{8B}} & \multirow{2}{*}{\texttt{128K}} & \hc{\texttt{79.43}} & \hc{\texttt{68.11}} & \hc{\texttt{25.69}} & \hc{\texttt{27.45}} & \hc{\texttt{25.46}} & \hc{\texttt{14.24}} \\
\addlinespace[-0.2em]
& & & & & \hc{\diffdo{3.04}} & \hc{\diffdo{4.83}} & \hc{\diffdo{3.90}} & \hc{\diffdo{3.17}} & \hc{\diffdo{9.35}} & \hc{\diffdo{2.37}} \\
\addlinespace[0.12em]
& \multirow{2}{*}{\textbf{\hsq{\texttt{OctoLong-8B-Instruct - Model Merging}}}} & \multirow{2}{*}{\gcode} & \multirow{2}{*}{\texttt{8B}} & \multirow{2}{*}{\texttt{128K}} & \hc{\texttt{83.08}} & \hc{\texttt{72.76}} & \hc{\texttt{28.22}} & \hc{\texttt{29.85}} & \hc{\texttt{35.08}} & \hc{\texttt{15.79}} \\
\addlinespace[-0.2em]
& & & & & \hc{\diffup{0.61}} & \hc{\diffdo{0.18}} & \hc{\diffdo{1.37}} & \hc{\diffdo{0.77}} & \hc{\diffup{0.27}} & \hc{\diffdo{0.82}} \\
\addlinespace[0.12em]
\bottomrule
\end{tabular}%
}
\caption{Detailed per-setting scores on the agentic tool-use evaluation suite (\textbf{\texttt{RQ3}}). \texttt{BFCL V3}~\citep{DBLP:conf/icml/PatilMYJSSG25} is split into its non-live, live, and multi-turn subsets, while \texttt{$\tau^2$-Bench}~\citep{DBLP:journals/corr/abs-2506-07982} reports its airline, retail, and telecom domains. Despite comparisons with several preference- and RL-tuned baselines, we observe consistently strong performance from the \texttt{OctoLong-Instruct} suite on short- and long-horizon generative tool-calling tasks. \cref{subsec:RQ3} details the results.}
\label{tab:Results_RQ3}
\end{table}
\begin{table}[H]
\centering
\scalebox{0.8}{\footnotesize \gcode\;\textbf{\texttt{Code-Specialised}} \qquad \gcext\;\textbf{\texttt{Context-Extended-After-Post-Training}} \qquad \gpost\;\textbf{\texttt{Preference/RL-Tuned}}}
\scalebox{0.55}{%
\setlength{\tabcolsep}{4pt}%
\begin{tabular}{@{}c l l r r rrrr rrrr@{}}
\toprule
& \multirow{3}{*}{\textbf{\texttt{Model}}} & & \multirow{3}{*}{\textbf{\texttt{Size}}} & \multicolumn{1}{c}{\multirow{3}{*}{\shortstack{\textbf{\texttt{Max}}\\ \textbf{\texttt{Length}}}}} & \multicolumn{4}{c}{\textbf{\texttt{LiveCodeBench V6}}} & \multicolumn{4}{c}{\textbf{\texttt{BigCodeBench Ins. Hard}}} \\
\cmidrule(lr){6-9} \cmidrule(lr){10-13}
& & & & & \hc{\textbf{\texttt{Pass}}} & \hc{\textbf{\texttt{Pass}}} & \hc{\textbf{\texttt{Pass}}} & \hc{\textbf{\texttt{Pass}}} & \hc{\textbf{\texttt{Pass}}} & \hc{\textbf{\texttt{Pass}}} & \hc{\textbf{\texttt{Pass}}} & \hc{\textbf{\texttt{Pass}}} \\
& & & & & \hc{\textbf{\texttt{@1}}} & \hc{\textbf{\texttt{@5}}} & \hc{\textbf{\texttt{@10}}} & \hc{\textbf{\texttt{@25}}} & \hc{\textbf{\texttt{@1}}} & \hc{\textbf{\texttt{@5}}} & \hc{\textbf{\texttt{@10}}} & \hc{\textbf{\texttt{@25}}} \\
\midrule
\multirow{1}{*}{\rotatebox[origin=c]{90}{\textbf{\texttt{XS}}}}
& \textbf{\hsq{\texttt{OctoLong-0.6B-Instruct}}} & \gcode & \texttt{0.6B} & \texttt{128K} & \olcg{4.55} & \olcg{9.24} & \olcg{11.26} & \olcg{13.86} & \olcg{1.60} & \olcg{5.10} & \olcg{7.92} & \olcg{14.80} \\
\midrule
\multirow{4}{*}{\rotatebox[origin=c]{90}{\textbf{\texttt{S}}}}
& \rmodel{meta-llama/Llama-3.2-1B-Instruct} & \gpost & \texttt{1B} & \texttt{128K} & \olcg{2.45} & \olcg{6.35} & \olcg{7.87} & \olcg{10.23} & \olcg{0.80} & \olcg{1.83} & \olcg{3.21} & \olcg{6.24} \\
& \rmodel{01-ai/Yi-Coder-1.5B-Chat} & \gcode\;\gpost & \texttt{1.5B} & \texttt{128K} & \olcg{8.91} & \olcg{15.61} & \olcg{17.34} & \olcg{20.87} & \olcg{5.11} & \olcg{15.99} & \olcg{20.88} & \olcg{27.45} \\
& \textbf{\hsq{\texttt{OctoLong-1.7B-Instruct}}} & \gcode & \texttt{1.7B} & \texttt{128K} & \olcg{10.79} & \olcg{18.34} & \olcg{20.94} & \olcg{24.53} & \olcg{4.90} & \olcg{16.83} & \olcg{23.61} & \olcg{32.34} \\
& \rmodel{ibm-granite/granite-3.1-2b-instruct} & \gpost & \texttt{2B} & \texttt{128K} & \olcg{9.57} & \olcg{14.36} & \olcg{17.43} & \olcg{19.78} & \olcg{6.21} & \olcg{13.24} & \olcg{18.06} & \olcg{23.96} \\
\midrule
\multirow{4}{*}{\rotatebox[origin=c]{90}{\textbf{\texttt{M}}}}
& \rmodel{meta-llama/Llama-3.2-3B-Instruct} & \gpost & \texttt{3B} & \texttt{128K} & \olcg{8.33} & \olcg{13.78} & \olcg{16.96} & \olcg{18.77} & \olcg{5.02} & \olcg{14.17} & \olcg{20.09} & \olcg{27.62} \\
& \rmodel{Qwen/Qwen3-4B-Instruct-2507} & \gpost & \texttt{4B} & \texttt{256K} & \olcg{17.86} & \olcg{29.44} & \olcg{33.56} & \olcg{36.76} & \olcg{19.87} & \olcg{28.12} & \olcg{30.76} & \olcg{33.46} \\
& \textbf{\hsq{\texttt{OctoLong-4B-Instruct}}} & \gcode & \texttt{4B} & \texttt{128K} & \olcg{17.44} & \olcg{27.91} & \olcg{31.34} & \olcg{34.91} & \olcg{16.31} & \olcg{31.70} & \olcg{39.59} & \olcg{44.82} \\
& \rmodel{arcee-ai/AFM-4.5B} & \gpost & \texttt{4.5B} & \texttt{64K} & \olcg{5.59} & \olcg{8.34} & \olcg{10.23} & \olcg{11.65} & \olcg{2.14} & \olcg{6.35} & \olcg{9.98} & \olcg{15.43} \\
\midrule
\multirow{12}{*}{\rotatebox[origin=c]{90}{\textbf{\texttt{L}}}}
& \rmodel{Qwen/Qwen2.5-7B-Instruct-1M} & \gpost & \texttt{7B} & \texttt{1024K} & \olcg{16.45} & \olcg{23.19} & \olcg{25.34} & \olcg{28.74} & \olcg{13.11} & \olcg{28.23} & \olcg{33.32} & \olcg{38.87} \\
& \rmodel{aws-prototyping/MegaBeam-Mistral-7B-512k} & \gcext & \texttt{7B} & \texttt{512K} & \olcg{2.77} & \olcg{5.84} & \olcg{6.87} & \olcg{9.41} & \olcg{0.52} & \olcg{0.96} & \olcg{1.56} & \olcg{3.12} \\
& \rmodel{internlm/internlm2_5-7b-chat-1m} & \gpost & \texttt{7B} & \texttt{1024K} & \olcg{2.82} & \olcg{4.95} & \olcg{6.54} & \olcg{10.23} & \olcg{4.23} & \olcg{14.34} & \olcg{18.83} & \olcg{24.97} \\
& \textbf{\hsq{\texttt{OctoLong-8B-Instruct}}} & \gcode & \texttt{8B} & \texttt{128K} & \olcg{20.82} & \olcg{28.78} & \olcg{33.86} & \olcg{37.24} & \olcg{19.44} & \olcg{37.16} & \olcg{42.84} & \olcg{49.23} \\
& \rmodel{meta-llama/Llama-3.1-8B-Instruct} & \gpost & \texttt{8B} & \texttt{128K} & \olcg{11.83} & \olcg{19.41} & \olcg{21.96} & \olcg{25.56} & \olcg{11.21} & \olcg{27.43} & \olcg{33.09} & \olcg{39.87} \\
& \rmodel{nvidia/Llama-3.1-Nemotron-8B-UltraLong-1M-Instruct} & \gcext & \texttt{8B} & \texttt{1024K} & \olcg{14.34} & \olcg{21.76} & \olcg{24.35} & \olcg{26.75} & \olcg{10.35} & \olcg{24.32} & \olcg{30.97} & \olcg{37.66} \\
& \rmodel{princeton-nlp/Llama-3-8B-ProLong-512k-Instruct} & \gcext & \texttt{8B} & \texttt{512K} & \olcg{5.12} & \olcg{9.45} & \olcg{11.49} & \olcg{14.29} & \olcg{7.21} & \olcg{18.34} & \olcg{23.32} & \olcg{28.76} \\
& \rmodel{ibm-granite/granite-3.1-8b-instruct} & \gpost & \texttt{8B} & \texttt{128K} & \olcg{12.87} & \olcg{17.95} & \olcg{21.45} & \olcg{25.44} & \olcg{13.12} & \olcg{26.45} & \olcg{32.87} & \olcg{38.45} \\
& \rmodel{gradientai/Llama-3-8B-Instruct-262k} & \gcext & \texttt{8B} & \texttt{256K} & \olcg{3.42} & \olcg{7.54} & \olcg{9.67} & \olcg{13.23} & \olcg{3.68} & \olcg{12.76} & \olcg{18.65} & \olcg{26.45} \\
& \rmodel{mistralai/Ministral-8B-Instruct-2410} & \gpost & \texttt{8B} & \texttt{128K} & \olcg{13.23} & \olcg{19.56} & \olcg{22.54} & \olcg{26.78} & \olcg{14.21} & \olcg{27.46} & \olcg{32.89} & \olcg{37.71} \\
& \rmodel{zai-org/glm-4-9b-chat-1m} & \gpost & \texttt{9B} & \texttt{1024K} & \olcg{13.62} & \olcg{19.54} & \olcg{22.04} & \olcg{25.47} & \olcg{11.33} & \olcg{27.22} & \olcg{33.01} & \olcg{39.54} \\
& \rmodel{01-ai/Yi-Coder-9B-Chat} & \gcode\;\gpost & \texttt{9B} & \texttt{128K} & \olcg{17.45} & \olcg{24.66} & \olcg{29.45} & \olcg{33.43} & \olcg{13.04} & \olcg{26.94} & \olcg{33.81} & \olcg{40.44} \\
\midrule
\multirow{2}{*}{\rotatebox[origin=c]{90}{\textbf{\texttt{XL}}}}
& \textbf{\hsq{\texttt{OctoLong-14B-Instruct}}} & \gcode & \texttt{14B} & \texttt{128K} & \olcg{22.43} & \olcg{32.77} & \olcg{35.86} & \olcg{41.77} & \olcg{21.52} & \olcg{38.43} & \olcg{44.19} & \olcg{51.26} \\
& \rmodel{Qwen/Qwen2.5-14B-Instruct-1M} & \gpost & \texttt{14B} & \texttt{1024K} & \olcg{20.89} & \olcg{29.23} & \olcg{33.42} & \olcg{35.68} & \olcg{19.61} & \olcg{33.67} & \olcg{37.65} & \olcg{43.54} \\
\midrule
\multirow{8}{*}{\rotatebox[origin=c]{90}{\textbf{\texttt{Abl.}}}}
& \multirow{2}{*}{\textbf{\hsq{\texttt{OctoLong-8B-Instruct - Cross-Repo Code}}}} & \multirow{2}{*}{\gcode} & \multirow{2}{*}{\texttt{8B}} & \multirow{2}{*}{\texttt{128K}} & \hc{\texttt{19.67}} & \hc{\texttt{26.57}} & \hc{\texttt{32.24}} & \hc{\texttt{35.91}} & \hc{\texttt{17.56}} & \hc{\texttt{35.04}} & \hc{\texttt{39.45}} & \hc{\texttt{43.56}} \\
\addlinespace[-0.2em]
& & & & & \hc{\diffdo{1.15}} & \hc{\diffdo{2.21}} & \hc{\diffdo{1.62}} & \hc{\diffdo{1.33}} & \hc{\diffdo{1.88}} & \hc{\diffdo{2.12}} & \hc{\diffdo{3.39}} & \hc{\diffdo{5.67}} \\
\addlinespace[0.12em]
& \multirow{2}{*}{\textbf{\hsq{\texttt{OctoLong-8B-Instruct - Model Merging}}}} & \multirow{2}{*}{\gcode} & \multirow{2}{*}{\texttt{8B}} & \multirow{2}{*}{\texttt{128K}} & \hc{\texttt{17.75}} & \hc{\texttt{24.56}} & \hc{\texttt{31.45}} & \hc{\texttt{34.59}} & \hc{\texttt{18.04}} & \hc{\texttt{36.35}} & \hc{\texttt{40.68}} & \hc{\texttt{46.05}} \\
\addlinespace[-0.2em]
& & & & & \hc{\diffdo{3.07}} & \hc{\diffdo{4.22}} & \hc{\diffdo{2.41}} & \hc{\diffdo{2.65}} & \hc{\diffdo{1.40}} & \hc{\diffdo{0.81}} & \hc{\diffdo{2.16}} & \hc{\diffdo{3.18}} \\
\addlinespace[0.12em]
& \multirow{2}{*}{\textbf{\hsq{\texttt{OctoLong-8B-Instruct - Context Extension}}}} & \multirow{2}{*}{\gcode} & \multirow{2}{*}{\texttt{8B}} & \multirow{2}{*}{\texttt{32K}} & \hc{\texttt{20.49}} & \hc{\texttt{28.91}} & \hc{\texttt{33.44}} & \hc{\texttt{36.95}} & \hc{\texttt{19.03}} & \hc{\texttt{36.55}} & \hc{\texttt{41.24}} & \hc{\texttt{47.98}} \\
\addlinespace[-0.2em]
& & & & & \hc{\diffdo{0.33}} & \hc{\diffup{0.13}} & \hc{\diffdo{0.42}} & \hc{\diffdo{0.29}} & \hc{\diffdo{0.41}} & \hc{\diffdo{0.61}} & \hc{\diffdo{1.60}} & \hc{\diffdo{1.25}} \\
\addlinespace[0.12em]
& \multirow{2}{*}{\textbf{\hsq{\texttt{Qwen3-8B}}}} & \multirow{2}{*}{\gpost} & \multirow{2}{*}{\texttt{8B}} & \multirow{2}{*}{\texttt{32K}} & \hc{\texttt{23.18}} & \hc{\texttt{29.31}} & \hc{\texttt{33.06}} & \hc{\texttt{34.91}} & \hc{\texttt{20.73}} & \hc{\texttt{34.46}} & \hc{\texttt{37.55}} & \hc{\texttt{40.11}} \\
\addlinespace[-0.2em]
& & & & & \hc{\diffup{2.36}} & \hc{\diffup{0.53}} & \hc{\diffdo{0.80}} & \hc{\diffdo{2.33}} & \hc{\diffup{1.29}} & \hc{\diffdo{2.70}} & \hc{\diffdo{5.29}} & \hc{\diffdo{9.12}} \\
\addlinespace[0.12em]
\bottomrule
\end{tabular}%
}
\caption{Detailed per-setting scores on the short-context code-generation evaluation suite (\textbf{\texttt{RQ4}}). \texttt{LiveCodeBench V6}~\citep{DBLP:conf/iclr/JainHGLYZWSSS25} and \texttt{BigCodeBench Instruct Hard}~\citep{DBLP:conf/iclr/ZhuoVCH0WYZHPB025} are reported at their \texttt{Pass@\{1,5,10,25\}} sampling budgets. We observe no regressions of the \texttt{OctoLong-Instruct} suite on short-context coding tasks along with important gains on API usage, particularly at higher sampling budgets. \Cref{subsec:RQ4} discusses the results in detail.}
\label{tab:Results_RQ4}
\end{table}
\endgroup

\end{document}